\documentclass[11pt]{article}

\usepackage[letterpaper, margin=1in]{geometry}
\usepackage[utf8]{inputenc}
\usepackage[T1]{fontenc}
\usepackage{lmodern}
\usepackage{hyperref}
\usepackage{url}
\usepackage{booktabs}
\usepackage{amsfonts}
\usepackage{nicefrac}
\usepackage{microtype}
\usepackage{xcolor}
\usepackage{graphicx}
\graphicspath{{figures/}{./}{figures_xfer/}{../../figures/}{../../}}
\usepackage{subcaption}
\usepackage{wrapfig}
\usepackage{amsmath}
\usepackage{amssymb}
\usepackage{mathtools}
\usepackage{natbib}
\usepackage{authblk}
\usepackage{algorithm}
\usepackage{algorithmic}
\usepackage[capitalize,noabbrev]{cleveref}

\hypersetup{
    colorlinks=true,
    linkcolor=blue!60!black,
    citecolor=blue!60!black,
    urlcolor=blue!60!black,
}

\title{Autonomous Continual Learning and Knowledge Transfer\\ via Cortex-Inspired Neural Assemblies}

\author[1]{Kevin L. McKee}

\date{}

\begin{document}

\maketitle

\begin{abstract}
Unlike the brain, artificial neural networks (ANNs) suffer from "catastrophic forgetting": complete overwriting of solutions when re-training on new functions ("concept shift") or on a disjoint domain of the current function ("domain shift").
ANNs require each batch of data to be roughly representative of the complete distribution, a typically expensive and impractical requirement.
In this paper, we introduce Functional Task Networks (FTNs), a method of continual learning and knowledge transfer modeled after several properties of the mammalian neocortex.
FTN is formulated as a distributed, stochastic mixture of experts in which a sparse, binary routing mask is computed at inference time using a small support sample of input-output pairs.
Per-batch reconfiguration of the router during training, together with mask-overlap-filtered replay, shapes a basis of overlapping, partially trained subnetworks.
Unrelated tasks route through mostly separate assemblies, related tasks overlap on shared components. 
On block-sequentially trained continual learning and knowledge transfer tasks FTN retains solutions nearly as well as disjoint-mask isolation and attains the lowest pre-update head-start loss among matched-architecture comparators. 
The learned mask-overlap geometry tracks the known concept-sharing structure where available. 
We show experimentally that under correlated concept shift, support-set mask reconfiguration is necessary even for retention. 
Under domain shift, the distributed bank with replay is sufficient to handle retention without the configurer. 
Forward transfer, or head-start, and backward transfer, are also contributed by support-set mask configuration.
\end{abstract}

\section{Introduction}\label{sec:intro}
Artifiical neural networks (ANNs) do not inherently include mechanisms to prevent their trained weights from being overwritten when the domain or target of training changes.
The term \textit{catastrophic forgetting} (CF) refers to the sudden loss of functionality due to such changes in the data distribution during training \citep{mccloskey1989catastrophic,french1999catastrophic}.
One way to prevent catastrophic forgetting is to isolate old, trained parameters to disjoint network subspaces, a class of methods called \textit{parameter isolation} \citep{mallya2018packnet,serra2018overcoming,wortsman2020supermasks}, akin to training separate networks on separate tasks.
But parameter isolation is at odds with another priority: \textit{knowledge transfer}, or how we re-use those networks to solve new tasks.
Supposing we have a way to perform knowledge transfer, how do we then continue to train the recycled combination of networks on a new task without overwriting their original respective functions?
The number of such questions quickly grows when considering a completely autonomous system, such as the human brain, over its lifespan in a naturalistically complex environment.

In this report, we introduce Functional Task Networks (FTN), a way of using ANNs based on established properties of the human neocortex to jointly and autonomously address parameter isolation and knowledge transfer.
FTN enables the continual learning of new tasks as flexible neural assemblies.
It is robust to changes in the input domain, or \textit{domain shift}, and changes in the distribution of target variables, or \textit{concept shift}, while providing a systematic approach to knowledge transfer.
Importantly, FTN autonomously addresses the complete suite of problems associated with continual learning, whereas contemporary comparisons widely tend to address only a subset of such questions and require manual interventions or supervision otherwise.

\paragraph{Concept and domain shift.}
Continual-learning benchmarks differ in \emph{what} changes between tasks. Under \emph{domain shift} the input distribution changes while the labeling rule remains broadly compatible. The most generic way to prevent forgetting under domain shift is shuffled replay-based methods \citep{lopez2017gradient,chaudhry2018efficient,buzzega2020dark}, in which data from previous tasks are mixed in training minibatches. The gradients over the tasks are averaged, so no single task dominates all of the weight updates. 
Under \emph{concept shift} the labeling rule or output function itself changes, so the same input must be mapped to different targets in different tasks. 
\Cref{fig:shifts} demonstrates both problems using a neural network in a real nonlinear regression experiment.

\paragraph{Block-sequential training.}
Block sequential training presents a maximally difficult regime for retention than i.i.d.\ minibatching: each long stretch of updates is dominated by one task at a time, intensifying forgetting on other tasks. 
It is also the schedule of most realistic continual deployments (sequential adapter installation, on-the-job fine-tuning, sliding-window streams).
The experiments here specifically target this training regime as a necessary part of the continual learning problem definition.
\Cref{fig:shifts} illustrates the relationship between block-sequential training and domain or concept shift.

\par\vfill
\noindent\begin{minipage}{\textwidth}
\captionsetup{type=figure}
\centering
\begin{subfigure}{\textwidth}
\centering
\includegraphics[width=0.78\linewidth]{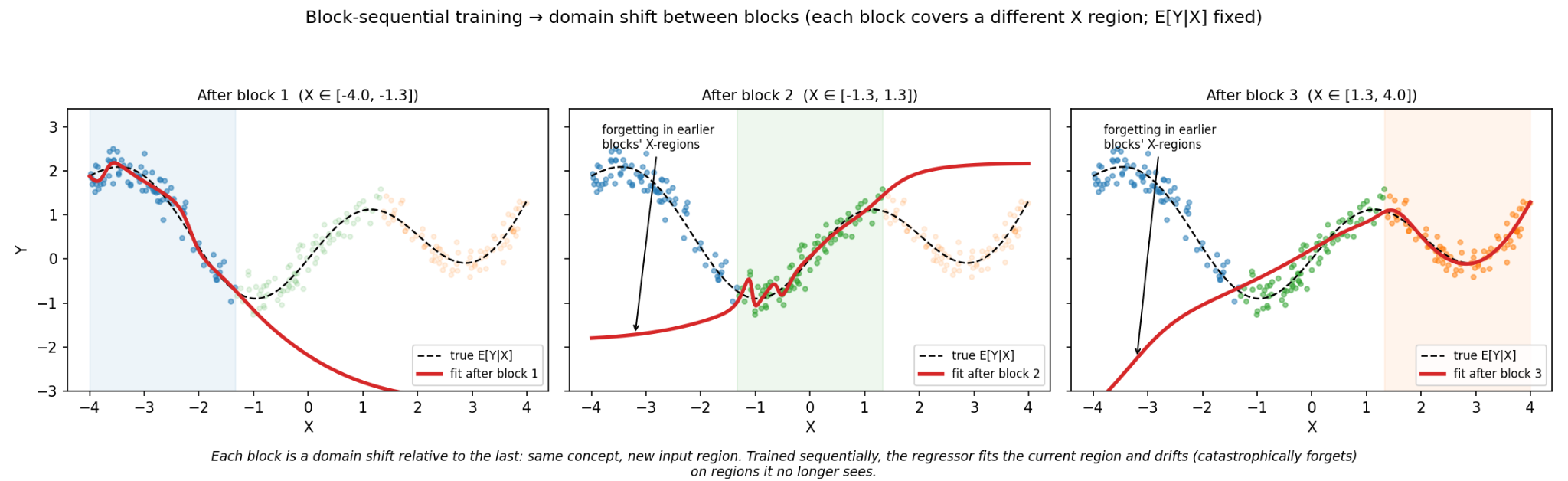}
\caption{Domain shift: new input regions, fixed target function.}
\label{fig:domain_shift}
\end{subfigure}

\smallskip
\begin{subfigure}{\textwidth}
\centering
\includegraphics[width=0.78\linewidth]{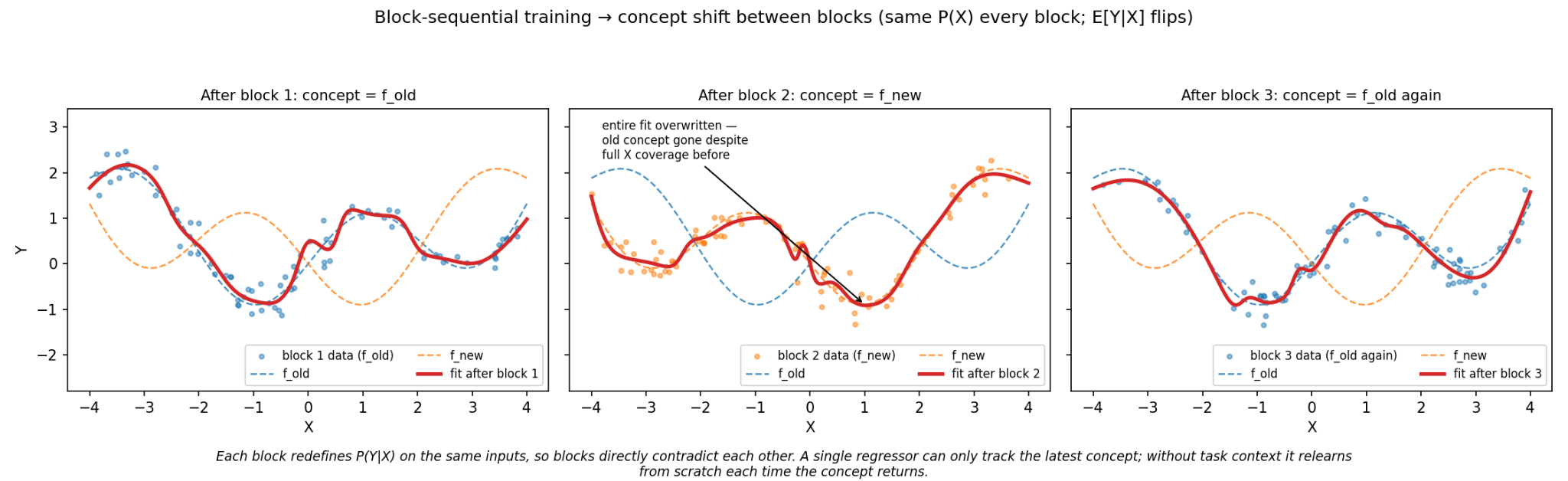}
\caption{Concept shift: new target function, fixed input distribution.}
\label{fig:concept_shift}
\end{subfigure}
\caption{Catastrophic forgetting under block-sequential nonlinear regression. Red curves show predictions after each block. Fitting a new input region degrades earlier regions; fitting a new target function overwrites the previous function.}
\label{fig:shifts}
\end{minipage}

\newpage

\subsection{Functional Task Networks.}

The principal idea behind FTN is to solve tasks using a \textbf{sparse, discrete} router over \textbf{large} bank of \textbf{small, trainable} features (or ``experts''), with the routing logits \textbf{directly optimized}, rather than computed from input.
Specifically, optimization of the router's logits is performed separately from optimization of the parameters of the processing units that are routed.
Each of these design choices above, emphasized in bold, are critical.

\begin{figure}[t]
\centering
\begin{subfigure}[b]{0.48\textwidth}
\centering
\includegraphics[width=\linewidth]{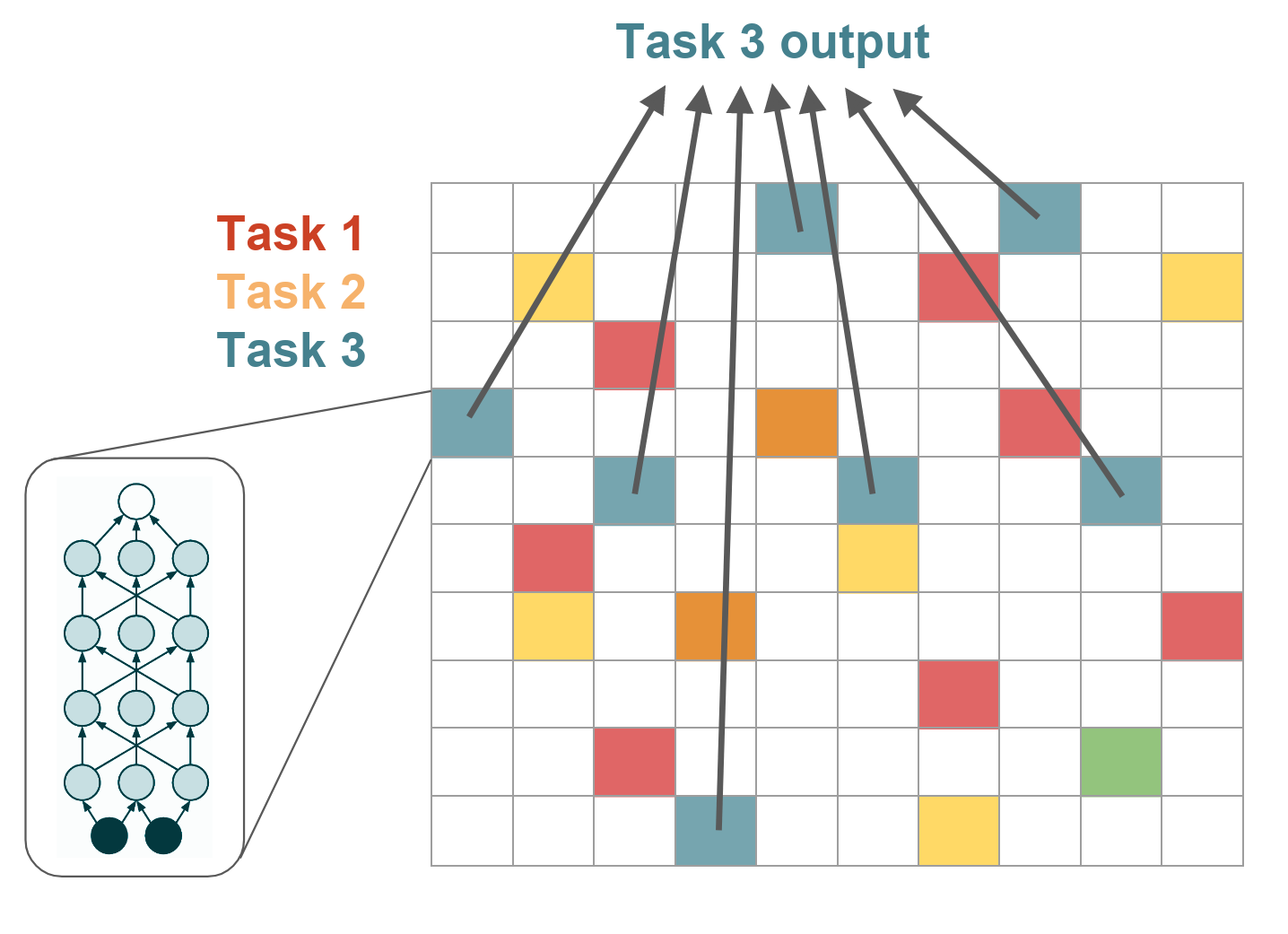}
\caption{Linear combination of $K$ gated MLPs}
\label{fig:model_base}
\end{subfigure}
\hfill
\begin{subfigure}[b]{0.48\textwidth}
\centering
\includegraphics[width=\linewidth]{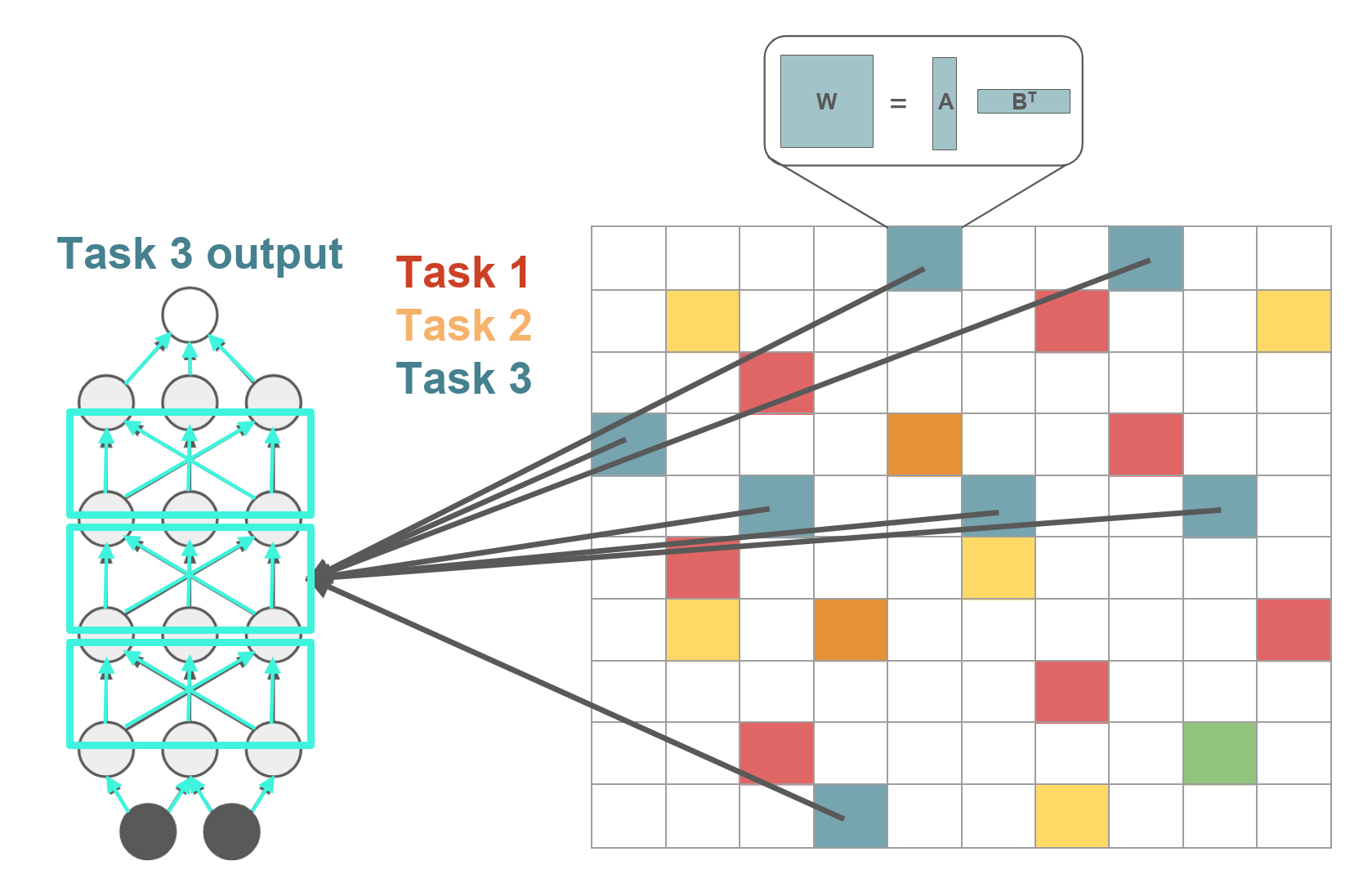}
\caption{Rank-$K$ recombinant mixture of LoRAs}
\label{fig:model_lora}
\end{subfigure}
\caption{Two possible architectures that implement continual learning via FTN. A large bank of processing units is discretely and sparsely gated into a combination function.}
\label{fig:archs}
\end{figure}

The implementation of FTN can vary to address the problem domain, but it necessarily involves a bank of processing units combined with an output decoder.
For instance, for simple motor policies, each unit can be a narrow-but-deep multi-layer perceptron (MLP) with a scalar output, and the output policy can be decoded via linear combination of the population activity, illustrated in \Cref{fig:model_base}.
Conversely, \Cref{fig:model_lora} shows how for Natural Language Processing (NLP) and other large-scale models, a sensible approach is to make each unit a Low-Rank Adapter (LoRA). The combination function is then the pretrained transformer such as a Large Language Model (LLM).

\paragraph{Optimization-based routing}
First, Mixture-of-Expert (MoE) models already route processing to model subspaces and can be made large, sparse, and discrete, but that does not solve catastrophic forgetting.
That is because the router's trained parameters are equally vulnerable to overwriting, the router itself being a trained function of the inputs. 

\begin{wrapfigure}{r}{0.35\textwidth}
\centering
\includegraphics[width=\linewidth]{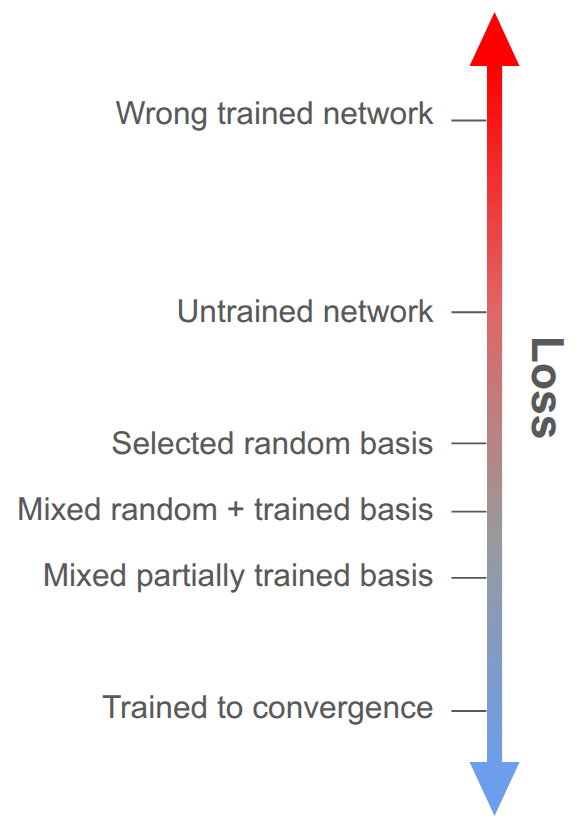}
\captionsetup{font=small}
\caption{Schematic loss ordering of candidate neural assemblies. Lower positions indicate lower loss on the target task. Relative performance depends on task structure and feature usefulness.}
\label{fig:loss}
\end{wrapfigure}
Routing via optimization allows the network to autonomously detect significant changes in loss on the current data stream and select an appropriate, alternative subspace in response.
Which subspace is appropriate: trained, untrained, or some combination, is addressed by the relative performance of those choices to one another in general.
\Cref{fig:loss} lines them up against a hypothetical loss scale.
A well-trained model will perform worse than a randomly initialized, untrained network on an unrelated problem.
An untrained model performs worse than a set of fixed random features selected via loss minimization, the principle behind random kernel expansion \citep{rahimi2007random} and \textit{reservoir computing} \citep{jaeger2001echo,lukosevicius2009reservoir}.
A mixture of partially trained and random features can be expected to perform better than random features alone, if tasks are structurally related and the features reflect the shared structure.
Finally, we expect a network trained to convergence to outperform any of the above on its training task.

When a new task appears to the network, gradients inherently perform parameter isolation by directing the routing mask away from a from a well-trained solution and toward the best-performing randomly intialized features.
If some previously trained features are more useful than random ones, they too can be recruited for the new task, providing an even larger head-start.
If a different but previously well-trained task appears, then the optimizer will select the routing solution corresponding to that trained network.  
Hence, new network instantiation, parameter isolation, task recall, and knowledge transfer are all controlled by the same mechanism.

For the above network-selection logic, the specific differences in performance among possible neural assemblies depends on the statistics of the tasks in question.
For deeply structured problems, such as image generation, selected random features and previously trained features may provide almost no benefit over random initial weights on a new task.
For instance, a model trained only on a variety of human faces does not learn the kinds of geometric forms needed to then generate architectural images.
In that case, the only reasonable expectation is nearly disjoint feature spaces, each trained to convergence.
Motor control presents a very different case: most tasks performed by a robotic arm share broad patterns of motor control and differ only in the final, precise trajectories of object manipulation.

\paragraph{Architecture}
The most important reason to specify a large bank of features is simply to have a sufficient learning capacity for future tasks.
However, a large bank of units also allows the same tasks to recruit and train a distribution of possible, composite solutions, a form of ensembling.
The experiments to follow will show that learning large, distributed task representations gives the network many more ways to assemble solutions on new tasks.

In addition to using a large network, limiting the complexity of the features being combined can force them to diversify their learned functions, creating more atomic, compositional, and therefore transferable solutions.
For instance, the processing units can be narrow-but-deep MLPs or rank-1 LoRAs \Cref{fig:model_lora}.

For continual learning to work without catastrophic forgetting, the router requires sparse, discrete routing masks.
Foremost, binarizing the routing logits found by the optimizer gates the flow of gradients into the trainable features.
Inactive units do not participate in the current task and their parameters remain fixed.

Routing sparsity controls the capacity of the network.
A simple sparsification is to set the top $k$ out of $N$ logits to one and all else to zero.
If all of the resulting networks are completely disjoint, the FTN can retain $N/k$ tasks.


\paragraph{Biological motivation.}
FTN is motivated by several features of neocortical computation that suggest a route to sparse, reusable, and partially overlapping task representations. Sparsification via top-$k$ masking has previously served as an abstraction for the effects of inhibitory competition among neurons\citep{yang2019task,cichon2015branch,mongillo2018inhibitory,rolls2010attractor,khona2022attractor,amari1977dynamics,maass2000computational,douglas2004neuronal}. Our choice to use a large number of deep networks with low-dimensional output is based on the dendritic model of pyramidal neurons, which can be characterized as deep, multilayer networks\citep{poirazi2003pyramidal,poirazi2020illuminating,beniaguev2021single}.
Alternatively, the LoRA formulation serves as a hypothesis about dendritic redundancy and compartmentalization, as many possible dendritic configurations can be selected via inhibitory networks. 

FTN also draws on a statistical property of cortical representations: task-relevant activity is distributed across populations and partially redundant, rather than localized to single reliable units \citep{georgopoulos1986neuronal,averbeck2006neural,cohen2011measuring}, while individual cortical neurons are variable across trials and only partially reliable \citep{tolhurst1983statistical,softky1993highly,shadlen1998variable,faisal2008noise,churchland2010stimulus,goris2014partitioning}. Algorithmically, this motivates using a large population of small subnetworks whose contributions are individually weak but heterogeneous, and whose training histories differ across tasks. A new task can then be configured by selecting an assembly from many partially trained units, allowing useful computation before any further weight update. In this view, transfer arises not from retrieving a single stored expert, but from recombining a distributed population of imperfect, task-relevant components.

\section{Related work}\label{sec:related}


The first class of related models focus primarily on retention of prior solutions. Parameter-isolation methods (PackNet \citep{mallya2018packnet}, HAT \citep{serra2018overcoming}, SupSup \citep{wortsman2020supermasks}) assign tasks to disjoint parameter subsets; regularizers (EWC \citep{kirkpatrick2017overcoming}, Synaptic Intelligence \citep{zenke2017continual}) penalize drift on parameters important for earlier tasks; replay methods (GEM/A-GEM \citep{lopez2017gradient,chaudhry2018efficient}, iCaRL \citep{rebuffi2017icarl}, DER++) use stored exemplars. These target retention; they have no explicit mechanism for forward transfer.
Furthermore, methods like Elastic Weight Consolidation (EWC) and Synaptic Intelligence \citep{kirkpatrick2017overcoming,zenke2017continual} address domain shift but not concept shift. 
Penalizing drift on parameters that \emph{need} to compute different functions on different tasks only blocks learning. 
Shuffled experience replay similarly fails because two replayed labels for the same input contradict each other unless replay is routed through different subnetworks.

Other methods target both retention and knowledge transfer. 
Progressive Networks \citep{rusu2016progressive} dedicate a fresh column per task with lateral connections to prior columns, attaining perfect retention at linear-in-tasks parameter cost. PathNet \citep{fernando2017pathnet} searches paths through a fixed module bank and reuses modules via path overlap, the closest conceptual analog to FTN. Modular networks with task-driven priors \citep{veniat2021mntdp} make module-reuse-vs-grow explicit; meta-learning approaches such as OML \citep{javed2019oml} target both axes through the meta-objective. We position FTN inside this family: a parallel-bank architecture with a binary support-set-derived mask that uses the same optimization to control isolation and transfer.

Among knowledge transfer methods, the most relevant to FTN are those that use inference-time support-set to determine the router state. 
Sparse expert models \citep{shazeer2017outrageously,fedus2022switch} and conditional channel-gated networks \citep{abati2020conditional} learn gates over subnetworks. More directly relevant, a family of methods uses a labeled support set to infer task identity or pick from a bank of stored task models: supermask retrieval in SupSup \citep{wortsman2020supermasks}, prompt-pool selection in L2P and DualPrompt \citep{wang2022learning,wang2022dualprompt}, and learned routers on support-set features. This is the central comparator class.

Finally, Top-$k$ routing requires optimization through a discrete constraint; we compare a sigmoid relaxation, an annealed straight-through estimator \citep{bengio2013estimating}, and a Gumbel-Sigmoid / Concrete relaxation \citep{jang2017categorical,maddison2017concrete} as three candidate methods for determining the routing mask.

\section{Method}\label{sec:method}

\paragraph{FTN-MLP.}
The basic version of FTN contains $H$ independent scalar-output MLPs $z_j=f_j(x)$ and a linear readout. For MLP $j$, with $L$ hidden layers, $h_{j,0}=x$, $h_{j,\ell+1}=\operatorname{ReLU}(W_{j,\ell}h_{j,\ell}+b_{j,\ell})$, and $z_j=\tanh(W_{j,L}h_{j,L}+b_{j,L})$. A binary mask $m\in\{0,1\}^H$ with $\lVert m\rVert_0=k$ gives
\[
  \hat{y}=W_{\mathrm{out}}\!\left(z\odot m\right),
  \qquad z=(z_1,\ldots,z_H).
\]
This model was configured to use $H=4096=64^2$, $L=8$, hidden width $8$, $k=256$. Masks are reconfigured every $4$ training batches with AnnealSTE for $S=30$ Adam steps at learning rate $0.3$ using $256$ support examples; replay uses a $1600$-entry buffer, stores $100$ examples per task, and draws $2$ filtered replay samples per batch.
We refer to this implementation as simply ``FTN'' from here on and use it for all subsequent experiments, diagnostics, and baseline comparisons.

\paragraph{FTN-LoRA.}
An alternative implementation can apply the FTN concept to a large bank of LoRAs to flexibly modify the weights of a fixed and possibly pretrained model.
FTN-LoRA uses one frozen random MLP backbone with trainable LoRA atom banks on every linear layer. For layer $q$, the masked forward map is
\[
  y_q=W_{q,0}x_q+\frac{\alpha}{k}\sum_{h=1}^{H}m_h
  \left(x_qA_{q,h}B_{q,h}+b_{q,h}\right),
\]
followed by ReLU between hidden layers; only the atom parameters $(A_{q,h},B_{q,h},b_{q,h})$ are trained. 
This model was configured to use the same $H=4096=64^2$, $k=256$, mask search, and replay settings as FTN-MLP, with a frozen $4$-hidden-layer width-$256$ MLP, rank $r=1$, $\alpha=256$, frozen-backbone weight scale $1$, nonzero $B$ and bias initialization scale $0.01$, per-atom bias enabled.

Besides an initial, preliminary result to replicate the performance of basic FTN on our synthetic tasks, we leave further examination and application of FTN-LoRA to subsequent work.

\paragraph{Mask reconfiguration}
A configurer $C$ receives the current model and a labeled support batch and returns a hard top-$k$ mask after $S$ inner steps with model weights fixed. We compare Sigmoid, AnnealSTE, and Gumbel-Sigmoid forms; AnnealSTE with $S{=}30$ was chosen as the default after performing sweeps across each of the three methods and values of $S$ (full details in \cref{app:impl}).

To train neurons shared across tasks, each step draws a $(x,y,m)$ triple from a $1600$-entry buffer filtered by mask overlap: a stored entry is admitted when its mask overlaps the current mask moderately, $\rho \in (0.2, 0.8]$. Entries with $\rho \le 0.2$ share too little mass to interfere, while entries with $\rho > 0.8$ are likely already trained by the current or proximal minibatches. Filtered replay therefore spends replay supervision only on the overlapping parameters, targeting task interference directly, while the disjoint complement is protected by mask gating. As a comparison, matched-memory uniform-random sampling (NoMask\,+\,ER in \cref{sec:results_mnist}) ablates the contribution of the filter.

\paragraph{Changepoint-driven reconfiguration}
As an alternative to reconfiguring at a fixed batch interval, we can allow sharp increases in training loss to trigger reconfiguration. We test this changepoint based method by configuring once on the first batch, then tracking an exponentially weighted mean and variance of the loss under the current mask and reconfigure when the loss exceeds the mean by five standard deviations (found by manual tuning). We used a smoothing factor of $0.9$ and suppressed triggers for ten batches after each reconfiguration while the loss statistics recover. Each reconfiguration uses AnnealSTE with $30$ Adam steps for FTN and $60$ for FTN-LoRA at learning rate $0.3$ and a support batch of $256$ examples.

\paragraph{FTN training and evaluation procedure.}
\Cref{alg:ftn} summarizes the full FTN training and evaluation loop, combining the configurer, the mask-overlap-filtered replay, and the support-set-driven evaluation protocol. The same configurer $C$ is used at every training step and at evaluation time; the only difference between training and evaluation is whether the model parameters $\theta$ receive a gradient step.

\begin{algorithm}[t]
\caption{Functional Task Networks: training and reconfig evaluation}
\label{alg:ftn}
\begin{algorithmic}[1]
\REQUIRE Tasks $\{\mathcal{T}_t\}_{t=1}^{T}$ arriving block-sequentially; bank of $H$ subnetworks $\{f_j\}$ with parameters $\theta$; readout $W_\mathrm{out}$; active budget $k$; configurer $C$ with $S$ inner steps; replay buffer $\mathcal{B}$ of capacity $N_\mathcal{B}$; replay overlap window $(\rho_\mathrm{lo}, \rho_\mathrm{hi}]$; support size $n_s$
\STATE Initialize $\theta$, $W_\mathrm{out}$, $\mathcal{B}\leftarrow\varnothing$
\FOR{task $t=1,\dots,T$}
    \STATE Reset optimizer state for $(\theta, W_\mathrm{out})$
    \FOR{each training minibatch $(x, y) \sim \mathcal{T}_t$}
        \STATE Draw support batch $(x_s, y_s)\sim \mathcal{T}_t$ of size $n_s$
        \STATE \textbf{// Mask configuration with $\theta$ frozen}
        \STATE Initialize mask logits $m_0$; \textbf{for} $s=1,\dots,S$ \textbf{do} $m_s \leftarrow$ configurer step $C(m_{s-1}; x_s, y_s, \theta)$
        \STATE $m_\mathrm{cur} \leftarrow \mathrm{TopK}(m_S, k) \in \{0,1\}^H$
        \STATE \textbf{// Mask-overlap-filtered replay}
        \STATE Sample $(x_r, y_r, m_r)$ uniformly from $\{(x', y', m')\in\mathcal{B} : (m'\!\cdot m_\mathrm{cur})/k \in (\rho_\mathrm{lo}, \rho_\mathrm{hi}]\}$
        \STATE \textbf{// Gradient step through the active mask, current plus replay}
        \STATE $\mathcal{L} \leftarrow \mathrm{loss}(W_\mathrm{out}(z(x)\odot m_\mathrm{cur}), y) + \mathrm{loss}(W_\mathrm{out}(z(x_r)\odot m_r), y_r)$
        \STATE Update $(\theta, W_\mathrm{out})$ with $\nabla \mathcal{L}$
        \STATE Optionally insert $(x, y, m_\mathrm{cur})$ into $\mathcal{B}$ (evict oldest if $|\mathcal{B}|=N_\mathcal{B}$)
    \ENDFOR
\ENDFOR
\STATE \textbf{Evaluation (reconfig protocol).} Given any target task $t^\star$ and an unlabeled query, draw a labeled support batch from $\mathcal{T}_{t^\star}$, run the configurer for $S$ inner steps with $\theta$ frozen to obtain a mask $m^\star$, and predict $\hat{y} = W_\mathrm{out}(z(x)\odot m^\star)$. Task identity is never disclosed.
\end{algorithmic}
\end{algorithm}

\section{Experiments}\label{sec:experiments}

\paragraph{Training and evaluation.}
In each of the following experiments, tasks arrive in contiguous blocks. Before every forward pass of FTN, its configurer selects a mask on the current minibatch, then the model updates through that mask only, with one mask-filtered replay sample drawn as described above. At evaluation, all models under comparison receive a small labeled support set from a target task, and for evaluating retention and transfer, a disjoint held-out batch from the same task. The task identity is never disclosed. The configurer selects a mask using the support set, and all metrics are computed on the held-out batch. For FTN, we refer to this as the \emph{reconfig} protocol. For diagnostics only, we also report a \emph{stored} protocol that reuses a saved training mask and therefore acts as a task-identity oracle.

\paragraph{Comparator suite.}
Seven comparators run on the same parallel-MLP backbone and support-set evaluation protocol. \emph{NoMask} (naive sequential fine-tuning), \emph{NoMask\,+\,ER} (uniform-random replay buffer matched in capacity to FTN's filtered buffer), and \emph{FixedMask} (deterministic disjoint per-task blocks, task identity at eval) are the full-sharing/replay/oracle-isolation baselines. \emph{SupSup} \citep{wortsman2020supermasks} freezes a randomly initialized parallel bank and learns a per-task score vector and readout, evaluated by support-set retrieval over stored masks. \emph{Progressive Networks} \citep{rusu2016progressive} use disjoint per-task $k$-blocks on the same bank with old columns frozen and lateral access from prior columns into the new readout, evaluated by support-set retrieval over the stored cumulative-column masks. \emph{FixedMask\,+\,Router} is the simpler PackNet-style retrieval baseline; \emph{FTN\,+\,Router} (FTN training, retrieval at eval) asks whether the FTN inner loop beats simple retrieval over the FTN-trained bank.

Several standard continual-learning methods, including EWC, SI, LwF, GEM/A-GEM, DER++, HAT, and PackNet, optimize primarily for retention under different architectural and evaluation assumptions than the support-conditioned routing setting studied here. In particular, many assume explicit task identity, fixed task boundaries, or architectures that do not naturally map onto FTN's large sparse-bank setting. Rather than presenting potentially misleading reimplementations under mismatched constraints, we construct architecture-matched controls that isolate the underlying mechanisms most relevant to FTN: replay-based full sharing (NoMask+ER), oracle parameter isolation (FixedMask), and support-set retrieval over isolated subnetworks (FixedMask+Router). PathNet, MNTDP, and L2P/DualPrompt require a non-MLP backbone and are deferred to the LoRA-bank instantiation. Per-benchmark mean$\pm$std tables are in \cref{app:revision_tables}.

\subsection{Benchmarks and metrics}\label{sec:setup}
\begin{figure}[!ht]
\centering
\includegraphics[width=0.95\textwidth]{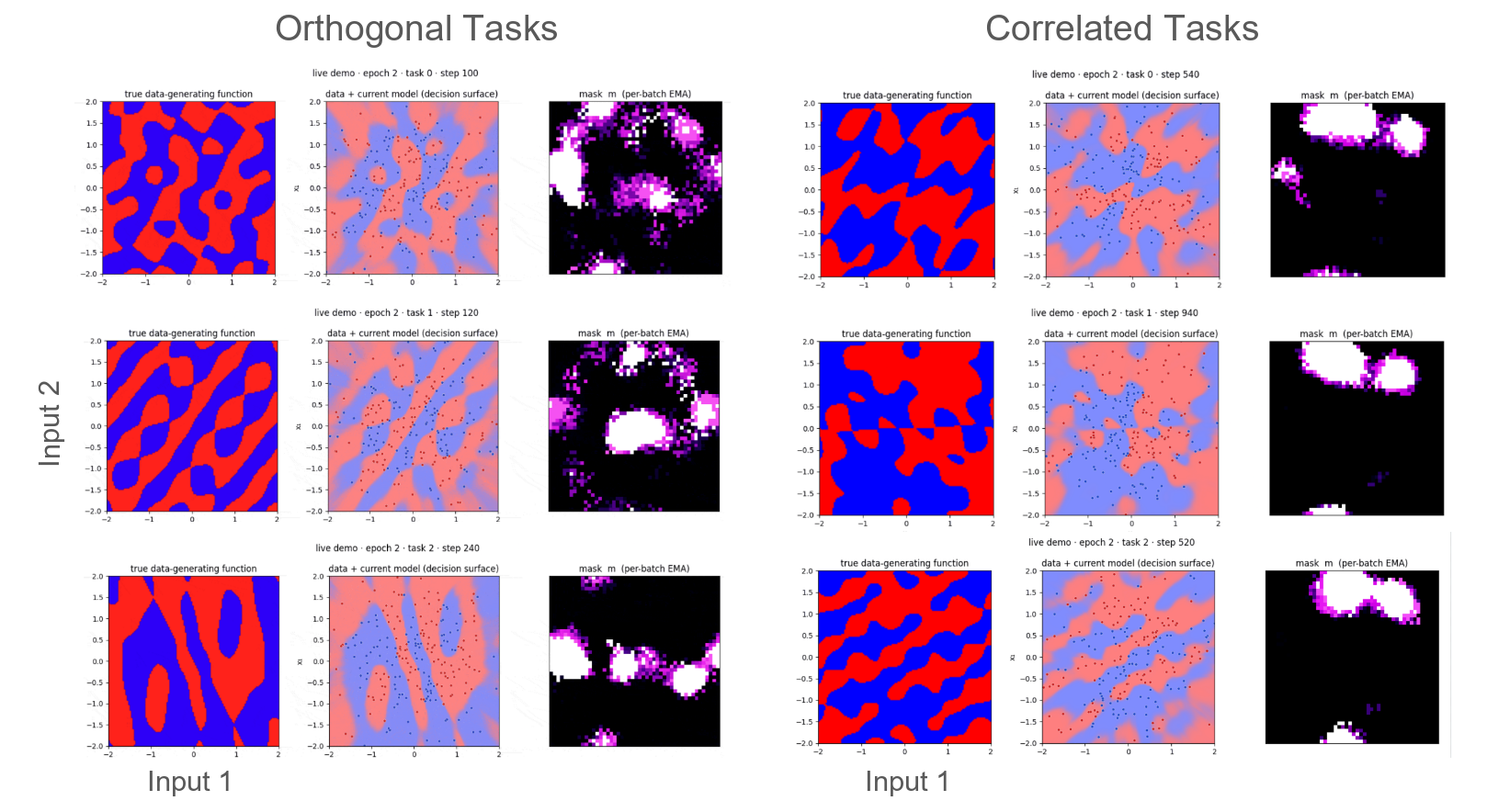}
\caption{Visualizations of randomly chosen batches, model expectations, and corresponding routing masks during training of FTN on the synthetic classification tasks.
Masks implemented a weak spatial cohesion bias for easier visual inspection of their structure.
Rows: Examples from each of 3 orthogonal (column 1) and 3 correlated (column 2) data-generating functions being trained in sequence. 
Each left-most pane shows the binary class-generating decision surface over 2 inputs.
Each middle pane shows a random batch of data during training superimposed on the model's estimate of that decision  surface.
The right-most pane shows the active routing mask (white pixels) superimposed on a colored exponential moving average (purple) of previous activations, show the distributed variation in the active mask.}
\label{fig:tasks}
\end{figure}

An class-label generating function of arbitrary difficulty for artificial neural networks can be generated by composing sinusoids of random input-conditioned frequencies. The \emph{synthetic} task family has $T{=}12$ binary classification tasks in $d{=}2$ generated from a fixed-encoder sinusoidal mixture over $C{=}6$ latent concept components: $x \sim \mathcal{N}(0, I_d)$, $\tilde z(x; t) = \sin\!\bigl(f\,(W_\mathrm{enc} x \odot m_t)\bigr)\odot m_t$, $y(x;t)=\mathbb{1}[\sum_l \tilde z_l > 0]$, with $f{=}8$ and $m_t\in\{0,1\}^L$ a binary task-component selector. The \emph{correlated} variant has $m_t$ select two of $C{=}6$ components (one of $\binom{6}{2}{=}15$ pairs, uniquely per task), so each component is shared by exactly two tasks; the \emph{orthogonal} variant uses $C{=}2T{=}24$ components partitioned into disjoint pairs (zero off-diagonal sharing). \Cref{fig:tasks} shows three example tasks mapping $(x_0, x_1)$ to $y$.

The MNIST suite~\citep{lecun1998mnist} spans concept and domain shift: \emph{shared-shuffle MNIST} (6 tasks) gives each task a digit-to-label permutation that is the union of two component sub-permutations, exposing ground-truth concept overlap; \emph{Permuted MNIST} and \emph{Rotated MNIST} (10 tasks each) are domain-incremental, applying a fixed random pixel permutation or a fixed rotation in equal steps from $0$ to $180^\circ$. Each MLP cell trains 1{,}000 Adam steps per task at lr $3{\times}10^{-4}$, train minibatch size $256$, and optimizer-state reset at task boundaries.

We report (1) accuracy (ACC) as the average classification accuracy across all tasks trained block-sequentially, (2) forgetting measure (FM) \citep{chaudhry2018efficient} i.e, the loss retroactively contributed by new tasks, and (3) backward transfer (BWT) \citep{lopez2017gradient} on the lower-triangular performance matrix under the reconfig protocol; (4) \emph{head-start loss} is the loss on a held-out batch from a new task after support-set mask configuration but before any weight update on that task, and lower values indicate stronger forward transfer. On the synthetic and shared-shuffle MNIST families, where ground-truth overlap is known, $r_{\mathrm{share}}$ is the Pearson correlation between off-diagonal entries of the row-normalized mask-overlap matrix and the ground-truth concept-sharing matrix.

\section{Results}

\subsection{Synthetic concept sharing}\label{sec:results_synthetic}

The correlated synthetic family is the only benchmark where ground-truth concept overlap is known by construction, so retention, forward transfer, and the recovered routing geometry can all be checked against a defined target. \Cref{tab:revision_synthetic} shows that unmasked sequential training collapses ($\mathrm{FM}{=}0.396$ for NoMask) and matched-memory random replay only partly recovers ($\mathrm{ACC}{=}0.65$). The four parameter-isolation methods reach $\mathrm{FM}{=}0$ to within sample noise (by construction for three; within $0.03$ for FTN despite no provided task identities), so retention does not separate the masked methods. The differences are on the other axes.

ACC reflects how much each isolation strategy lets the model learn: SupSup at $0.768$ is bottlenecked by its frozen random backbone, Progressive and FixedMask reach the oracle's $\approx 0.95$ by allocating fresh trained capacity per task, and FTN sits a few points below ($0.92$) because its bank weights are shared rather than partitioned.

Head-start loss measures how well the active subnetwork is already configured for a new task before any weight update (averaged across tasks in the sequence): FTN at $0.40$ is below every other comparator, including Progressive ($1.42$, ${\sim}3.5\times$) and SupSup ($0.66$).
Interestingly, FTN\,+\,Router reproduces FTN's number ($0.394$ vs.\ $0.399$) while the same retrieval on top of FixedMask, SupSup, or Progressive cannot.
For the router version, which retrieved masks from an explicit memory, the per-batch reconfiguration during \emph{training} is what populates that memory with a large variety of partly-trained overlapping subnetworks. It follows from these results that taking the $\arg\min$ of the loss over that bank is a viable but much more computationally expensive alternative to the reconfiguration's gradient-based search, which only used 30 steps.

The no-replay ablation raises FM about seven-fold (to $0.195$) and drops ACC to $0.78$, which localizes retention to the filtered-replay channel that protects shared neurons. Random per-batch masks (FTN training pipeline, configurer replaced by a fresh random $k$-of-$H$ mask each step) retain and transfer substantially worse than FTN under correlated concept shift, with ACC dropping from $0.919$ to $0.774$, FM rising from $0.029$ to $0.119$, and head-start loss rising from $0.399$ to $1.194$, showing that stochastic sparse-bank training plus replay is insufficient when tasks share latent components and that the support-conditioned configurer is necessary to route around the resulting cross-task mask collisions.

\begin{table}[t]
\centering
\small
\caption{Correlated synthetic family (12 tasks) under the reconfig (no-task-identity) protocol. Mean$\pm$std over 5 seeds. Higher ACC and BWT, lower FM and head-start (HS) loss, are better.}
\label{tab:revision_synthetic}
\begin{tabular}{lllll}
\toprule
Method & ACC$\uparrow$ & FM$\downarrow$ & BWT$\uparrow$ & HS$\downarrow$ \\
\midrule
NoMask (naive seq.) & $0.607 \pm 0.019$ & $0.396 \pm 0.017$ & $-0.396 \pm 0.017$ & $2.868 \pm 0.339$ \\
NoMask + ER (random replay) & $0.651 \pm 0.029$ & $0.165 \pm 0.011$ & $-0.164 \pm 0.010$ & $1.074 \pm 0.081$ \\
FixedMask (oracle ID at eval) & $0.948 \pm 0.022$ & $0.000 \pm 0.000$ & $0.000 \pm 0.000$ & $0.702 \pm 0.003$ \\
FixedMask + retrieval router & $0.948 \pm 0.021$ & $0.000 \pm 0.000$ & $0.000 \pm 0.000$ & $0.702 \pm 0.004$ \\
FTN (ours) & $0.919 \pm 0.020$ & $0.029 \pm 0.007$ & $-0.021 \pm 0.015$ & $0.399 \pm 0.064$ \\
FTN, random masks (ablation) & $0.774 \pm 0.089$ & $0.119 \pm 0.042$ & $-0.113 \pm 0.045$ & $1.194 \pm 0.198$ \\
FTN + retrieval router & $0.904 \pm 0.028$ & $0.045 \pm 0.008$ & $-0.043 \pm 0.011$ & $0.394 \pm 0.078$ \\
FTN, no replay & $0.780 \pm 0.036$ & $0.195 \pm 0.028$ & $-0.192 \pm 0.031$ & $0.521 \pm 0.064$ \\
\bottomrule
\end{tabular}
\end{table}

\paragraph{16-task retention and transfer}\label{null_vs_full}

\begin{figure}[!ht]
\centering
\begin{subfigure}[t]{0.32\textwidth}
\includegraphics[width=\textwidth]{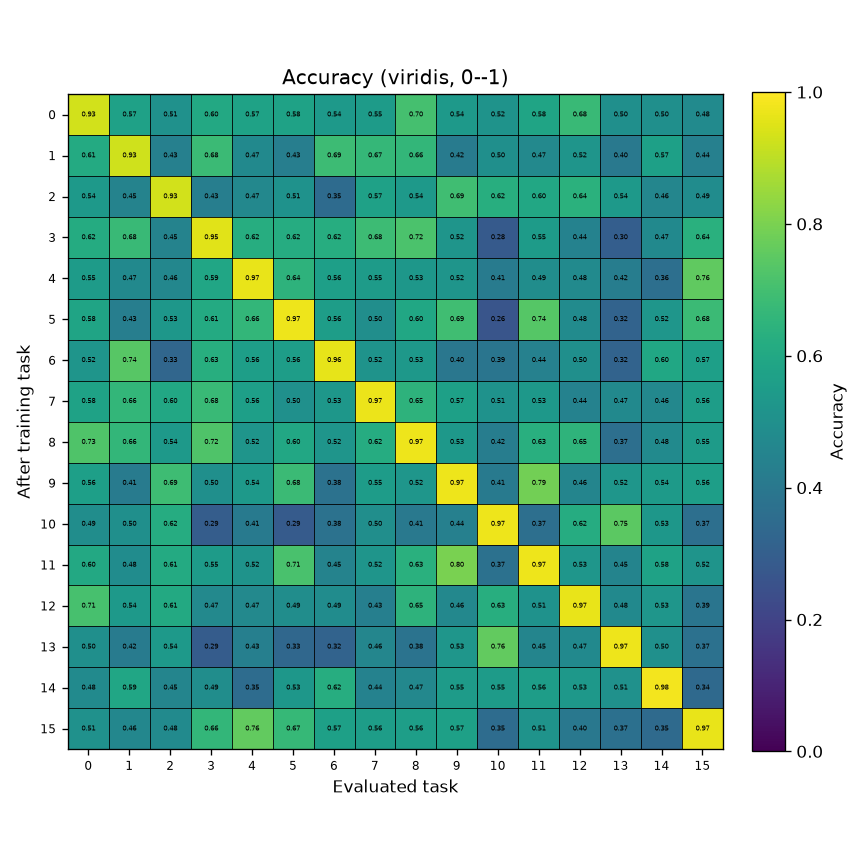}
\caption{Null result with the full network capacity and replay but no masking.}
\label{fig:null}
\end{subfigure}\hfill
\begin{subfigure}[t]{0.32\textwidth}
\includegraphics[width=\textwidth]{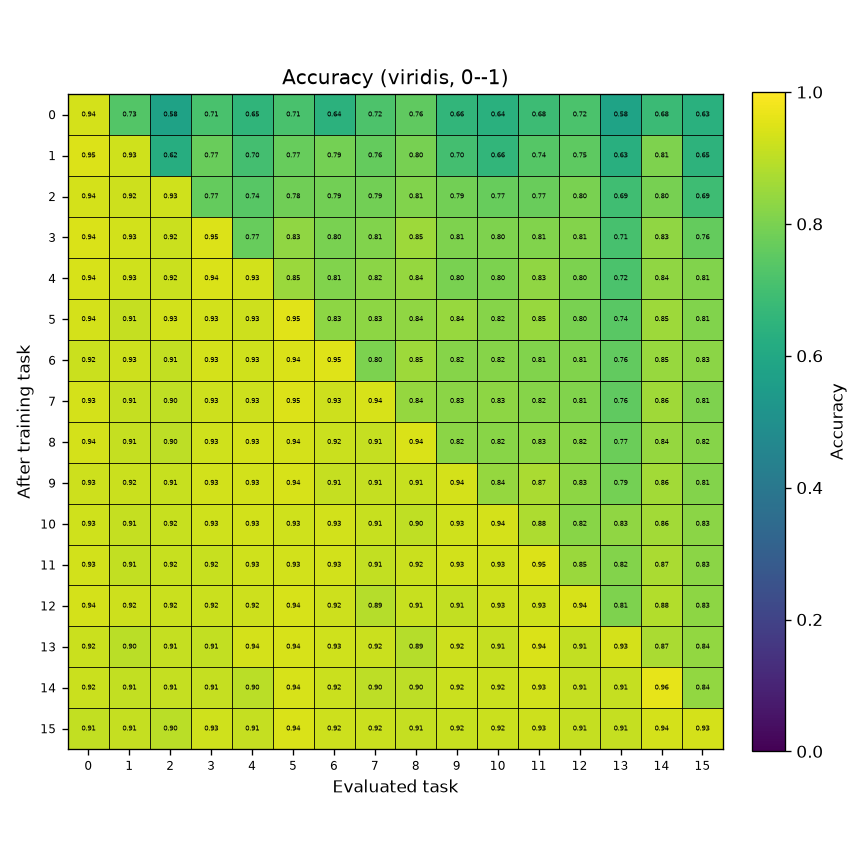}
\caption{FTN linear routed mixture of MLPs.}
\label{fig:full}
\end{subfigure}
\hfill
\begin{subfigure}[t]{0.32\textwidth}
\includegraphics[width=\textwidth]{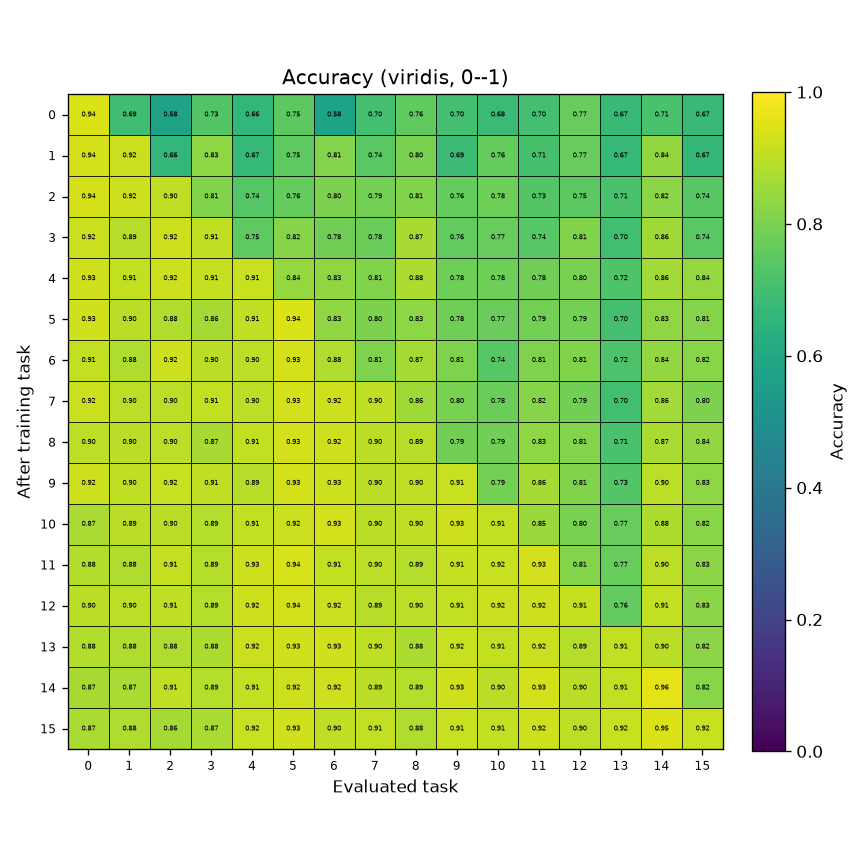}
\caption{FTN routed mixture of LoRAs applied to one large MLP.}
\label{fig:lora}
\end{subfigure}
\caption{Per-task evaluation matrices on the $16$-task synthetic sequence, same seed and same pipeline evaluating a generic model versus two implementations of FTN. Rows index training task; columns index evaluation task. The null illustrates expected behavior of generic MLPs under block-sequential training; the FTN implementations recover retention and forward transfer on the same task family.}
\label{fig:null_vs_full}
\end{figure}

\Cref{fig:null_vs_full} contrasts a generic MLP with experience replay with two FTN implementations on the same task family, same seed, and otherwise identical pipeline. All three panels are $16\times 16$ task-vs-task evaluation matrices obtained by training each model sequentially on $16$ tasks from the synthetic data generator and then evaluating the resulting model on each task's held-out batch at every intermediate checkpoint.
\Cref{fig:null} is the \emph{null} result, with no masking and full network capacity. Only the diagonal displays high accuracy: the model performs well its most recent task, with no retention (low scores in the lower triangle) and no forward transfer (low scores in the upper triangle). \Cref{fig:full} shows the FTN linear routed mixture of MLPs, while \Cref{fig:lora} shows the FTN routed mixture of LoRAs applied to one large MLP. Both FTN implementations recover substantially better retention and forward transfer than the null result.

In \Cref{fig:null_vs_full}, the null result shows that as the MLP trains on each task, held-out correlated tasks get a small to moderate boost in performance, but
both current and hold-out task performance disappears as soon as the network retrains on the next task.
By contrast, both FTN models retain the current task and maintain their head start on held-out tasks after retraining on subsequent tasks. Where task-invariant representations did not exist, the distributed and stochastic training of the network produced a large repertoire of random and partially-trained task bases that nonetheless allowed rapid construction of solutions on each new task.

\paragraph{Replay-free, changepoint-driven routing.}
In a previous iteration of this work, we show that tasks can be retained and recalled effectively without experience replay when reconfiguration is triggered only at task boundaries \citep{mckee2026cortex}.
The current version has some design differences, including the annealed straight-through estimator and no spatial consolidation bias. 
We therefore tested the version of FTN presented here but with filtered experience replay disabled and with reconfiguration triggered only by sharp changes in loss, which was expected at task boundaries. 

We compared the MLP and LoRA implementations across fixed-interval and changepoint-driven mask reconfiguration, each with filtered replay or no replay. In this single-seed, 16-task run, all runs without replay performed worse overall on average eval accuracy for current and previous tasks than runs that used replay.
However, switching the replay-free models to changepoint-driven reconfiguration (i.e., reconfiguring once per task rather than every four batches), recovered $0.081$ points of accuracy in the base FTN model and $0.230$ in the LoRA model.
Both changepoint-driven reconfiguration and disabling replay reduced average mask overlap and forward transfer, supporting the hypothesis that more distributed task representations led to more transferable learning.

\begin{table}[!ht]
\centering
\small
\caption{Recorded continual-learning metrics on the 16-task sequence. Each condition is one seed. ``Fixed'' denotes fixed-interval mask reconfiguration; Reactive-Low uses loss-triggered reconfiguration. Recall and forward-transfer accuracy are averaged over the corresponding task evaluations; forward-transfer change is the mean change per task block.}
\label{tab:reactive_replay_metrics}
\scriptsize
\setlength{\tabcolsep}{3pt}
\begin{tabular}{lllrrrrrr}
\toprule
Model & Routing & Replay & Recall$\uparrow$ & Forward$\uparrow$ & FT change$\uparrow$ & Train$\uparrow$ & Current$\uparrow$ & Overlap (\%) \\
\midrule
MLP & Fixed & Filtered & 0.922 & 0.785 & 0.0118 & 0.950 & 0.941 & 15.78 \\
MLP & Fixed & None & 0.805 & 0.735 & 0.0055 & 0.954 & 0.946 & 14.76 \\
MLP & Changepoint-Driven & Filtered & 0.912 & 0.771 & 0.0107 & 0.952 & 0.923 & 13.45 \\
MLP & Changepoint-Driven & None & 0.886 & 0.756 & 0.0111 & 0.947 & 0.929 & 12.40 \\
\hline
LoRA & Fixed & Filtered & 0.906 & 0.780 & 0.0093 & 0.908 & 0.916 & 16.93 \\
LoRA & Fixed & None & 0.656 & 0.663 & 0.0012 & 0.825 & 0.853 & 40.38 \\
LoRA & Changepoint-Driven & Filtered & 0.911 & 0.776 & 0.0070 & 0.937 & 0.920 & 10.03 \\
LoRA & Changepoint-Driven & None & 0.890 & 0.775 & 0.0078 & 0.936 & 0.917 & 7.89 \\
\bottomrule
\end{tabular}
\end{table}

A more thorough comparison of the two strategies will require more seeds to allow statistical testing of differences. 
On this seed, the differences were small but accorded with our expectations given the hypothesized roles of distributed representation and replay.
Replay allows direct training of cross-task representations by mixing gradients from different task losses. Distributed representation, i.e. overlapping, redundant, and varied routing masks achieved by frequent within-task reconfiguration, should produce an ensembling effect: multiple ways to construct assemble a solution from the pool of trainable units.

\paragraph{Sparsity, support size, and the orthogonal control.}
Three diagnostic sweeps are summarized here; full tables and figures are in \cref{app:ablation_tables,app:sec_exp}. (i)~The top-$k$ budget controls how much overlap is possible: sweeping $k\in\{512,128,32,8\}$ reveals an isolation-transfer trade-off. FM falls from $0.056$ at $k{=}512$ to $0.032$ at $k{=}32$ while head-start rises from ${\approx}0.42$ to ${\approx}0.60$; sharing fidelity peaks at moderate sparsity ($r_{\mathrm{share}}{=}0.624$ at $k{=}32$). (ii)~Support-set size affects transfer more than retention: shrinking the support from $256$ to $4$ raises head-start from $0.425{\pm}0.080$ to $1.339{\pm}0.095$ while FM stays at or below $0.060$ and ACC at $0.86$ to $0.89$. (iii)~The orthogonal variant, which removes latent sharing while preserving the task format, raises FTN head-start from $0.423{\pm}0.077$ to $0.568{\pm}0.045$ with low forgetting in both regimes; $r_{\mathrm{share}}$ is zero by construction for the orthogonal task family.

\begin{figure}[!b]
\centering
\begin{subfigure}[b]{0.4\textwidth}
\includegraphics[width=\textwidth]{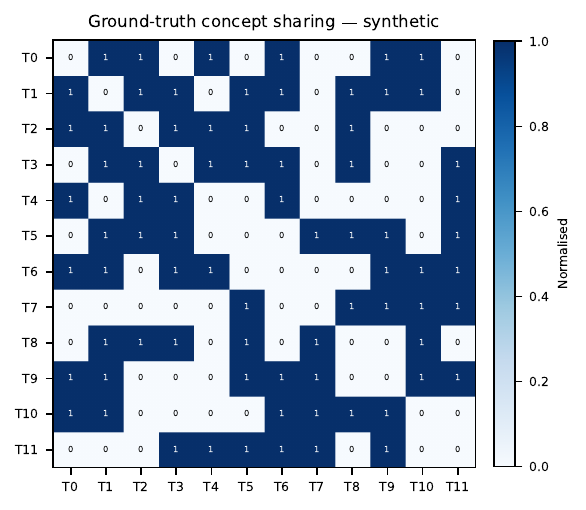}
\caption{Ground-truth.}
\end{subfigure}\hfill
\begin{subfigure}[b]{0.4\textwidth}
\includegraphics[width=\textwidth]{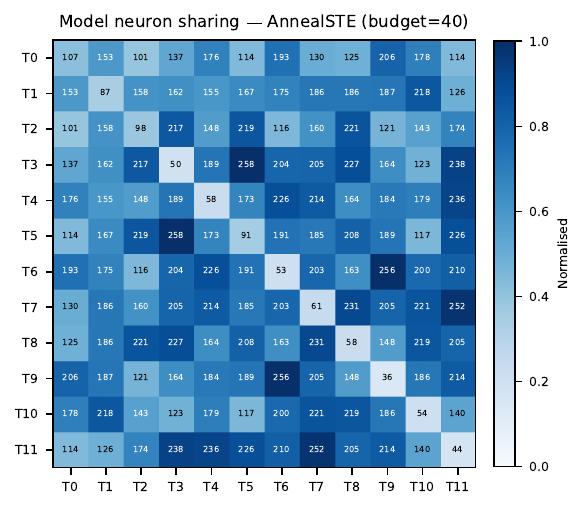}
\caption{Model, AnnealSTE $S{=}40$, $r_{\mathrm{share}}{=}0.617$.}
\end{subfigure}
\caption{Recovered task structure on the correlated synthetic family. Off-diagonal entries $S[i,j]$ count neurons (left) or concept components (right) shared between tasks $i\ne j$; the diagonal counts neurons / components used only by task $i$. $r_{\mathrm{share}}$ is computed on row-normalized off-diagonal entries.}
\label{fig:sharing_and_bwt}
\end{figure}

\paragraph{Recovered sharing geometry.}
\Cref{fig:sharing_and_bwt} compares the ground-truth sharing matrix with the row-normalized neuron-sharing matrix for a high-fidelity AnnealSTE configuration ($S{=}40$, $r_{\mathrm{share}}{=}0.617$; the primary sweep reaches $0.622$). The recovered matrix is not binary, but its high-overlap regions align with the concept-sharing pattern. BWT stays close to zero across the 18 primary-search configurations, ranging from $-0.061$ to $+0.005$, against $-0.392$ for NoMask on synthetic and $-0.656$ on MNIST.

\subsection{MNIST concept and domain shift}\label{sec:results_mnist}
MNIST provides a non-synthetic validation that the transfer pattern is not an artifact of the sinusoidal generator. \Cref{tab:mnist_suite} compresses the three variants. On shared-shuffle MNIST (concept shift), random replay fails as predicted, NoMask and NoMask+ER collapse relative to the masked methods and have severe forgetting, while FTN retains within $0.023$ FM and has much lower head-start loss than all isolation/retrieval baselines ($0.562$ vs.\ $2.309$ for FixedMask and $8.330$ for Progressive Networks; Progressive's lateral inputs carry the old digit-to-label map, which actively conflicts with the new permutation). On Permuted MNIST (non-smooth domain shift), replay and isolation already retain well, but FTN still has the lowest head-start loss. On Rotated MNIST (smooth domain shift), Progressive comes close ($0.381$ vs.\ FTN's $0.324$), the expected case where smooth lateral transfer is useful. Across all three MNIST families, FTN gives up a small amount of final accuracy to Progressive/FixedMask but has the lowest head-start loss among the matched-backbone comparators. The FTN-with-random-per-batch-masks ablation row holds the FTN training pipeline fixed (filtered replay, the same step budget, same parameter budget) and replaces the support-conditioned configurer with a fresh random $k$-of-$H$ mask each step. This ablation retains nearly as well as FTN on all three MNIST variants ($0.92$/$0.92$/$0.87$ ACC) but pays a head-start cost that depends on input-shift type ($\sim\!12\times$ worse on shared-shuffle MNIST, $6.65$ vs.\ $0.56$; $\sim\!2\times$ worse on Permuted, $3.41$ vs.\ $1.62$; and Rotated, $0.62$ vs.\ $0.32$). FTN+Router and no-replay diagnostics, together with the per-benchmark mean$\pm$std tables, are in \cref{app:revision_tables}.

\begin{table}[t]
\centering
\scriptsize
\setlength{\tabcolsep}{2.5pt}
\caption{Compressed MNIST suite under the reconfig protocol. Means over 5 seeds; full mean$\pm$std tables are in \cref{app:revision_tables}. Higher ACC and lower FM/HS are better.}
\resizebox{\textwidth}{!}{%
\begin{tabular}{lccccccccc}
\toprule
& \multicolumn{3}{c}{Shared-shuffle} & \multicolumn{3}{c}{Permuted} & \multicolumn{3}{c}{Rotated} \\
\cmidrule(lr){2-4}\cmidrule(lr){5-7}\cmidrule(lr){8-10}
Method & ACC$\uparrow$ & FM$\downarrow$ & HS$\downarrow$ & ACC$\uparrow$ & FM$\downarrow$ & HS$\downarrow$ & ACC$\uparrow$ & FM$\downarrow$ & HS$\downarrow$ \\
\midrule
NoMask & 0.444 & 0.644 & 10.397 & 0.928 & 0.053 & 5.207 & 0.523 & 0.509 & 0.360 \\
NoMask+ER & 0.442 & 0.630 & 8.524 & 0.960 & 0.018 & 5.270 & 0.912 & 0.075 & 0.667 \\
SupSup & 0.886 & 0.000 & 3.535 & 0.894 & 0.000 & 2.683 & 0.894 & 0.000 & 0.861 \\
Progressive Nets & 0.979 & 0.000 & 8.330 & 0.977 & 0.000 & 3.096 & 0.977 & 0.000 & 0.381 \\
FixedMask & 0.979 & 0.000 & 2.309 & 0.979 & 0.000 & 2.321 & 0.979 & 0.000 & 2.322 \\
FixedMask+Router & 0.978 & 0.000 & 2.311 & 0.979 & 0.000 & 2.320 & 0.979 & 0.000 & 2.321 \\
FTN, random per-batch masks (ablation) & 0.923 & 0.027 & 6.652 & 0.915 & 0.017 & 3.411 & 0.867 & 0.087 & 0.621 \\
FTN (ours) & 0.960 & 0.023 & \textbf{0.562} & 0.954 & 0.024 & \textbf{1.616} & 0.941 & 0.043 & \textbf{0.324} \\
\bottomrule
\end{tabular}%
}
\label{tab:mnist_suite}
\end{table}

\section{Discussion}\label{sec:discussion}
In this paper we presented a brain-inspired, unified, and autonomous approach to continual learning and knowledge transfer. We compared the novel method on both synthetic tasks to conduct diagnostics and on applied tasks to examine generality. We find that FTN, as a general design principle of autonomous, distributed representation, can be effectively implemented in at least two ways: either as a routed linear combination of small but deep MLPs, or as a population of recombinant LoRAs applied to a fixed model.

Both methods result in (1) retention and recovery of previously trained solutions when either the domain or the target changes, (2) transfer of training improvements across tasks both to held-out tasks and backward to previously trained tasks, (3) successful autonomous determination of when and which assemblies to recruit such that the above benefits are realized.
By comparison, regularization methods such as EWC and Synaptic Intelligence primarily protect parameters important to earlier tasks and are not designed for concept shift, while replay methods such as GEM/A-GEM, iCaRL, and DER++ primarily target retention and can encounter contradictory labels when the target function changes \citep{kirkpatrick2017overcoming,zenke2017continual,lopez2017gradient,chaudhry2018efficient,rebuffi2017icarl,buzzega2020dark}. Parameter-isolation methods such as PackNet, HAT, and SupSup protect prior solutions by restricting which parameters can be trained, but do not by themselves provide FTN's flexible recombination mechanism \citep{mallya2018packnet,serra2018overcoming,wortsman2020supermasks}. Progressive Networks and related modular methods add transfer through lateral connections, path reuse, or task-driven module selection, but use task-specific columns, paths, or stored task structure rather than one shared support-conditioned mechanism \citep{rusu2016progressive,fernando2017pathnet,veniat2021mntdp}.
Finally, many routing and retrieval systems require task-specific supervision or additional selection machinery: FixedMask and Progressive Networks use task-specific allocation, while SupSup, L2P, and DualPrompt retrieve task-associated masks, prompts, or modules \citep{wortsman2020supermasks,rusu2016progressive,wang2022learning,wang2022dualprompt}. FTN combines parameter protection, transfer through overlapping assemblies, and support-conditioned selection in one procedure, without exposing task identity or requiring manual intervention at task boundaries.
FTN was designed to meet all three and be deployable as a ``hands-off'' system, making minimal assumptions about the data and requiring only the data stream itself.

\paragraph{Comparison to other parameter isolation baselines.}
The ablative retrieval routers play two diagnostic roles. FixedMask\,+\,Router (parameter-isolation training, $\arg\min$-loss retrieval at eval) matches the task-identity oracle on retention but pays a high head-start cost because every retrieved block is fixed and untrained for a new task. FTN\,+\,Router (FTN training, retrieval at eval) inherits FTN's training-time benefit and produces a head-start loss comparable to FTN itself, so the head-start advantage stems from training-time per-batch reconfiguration that builds a broad basis of partially trained subnetworks rather than from the inner-loop optimization at evaluation. The head-start gap between FTN and FixedMask\,+\,Router is the central empirical signal, and it is largest on benchmarks whose tasks share transferable structure and smallest where the input shift carries the least such structure (Permuted MNIST). SupSup, Progressive, and FixedMask each pay a different transfer cost: SupSup is capped by its frozen random features; Progressive's columns are task-specialized, so its head-start is useful only when smoothly-parameterized lateral transfer happens to apply (helpful on Rotated MNIST, counterproductive on shared-shuffle); FTN's single shared bank with overlapping training masks delivers the lowest head-start on every benchmark. The cost is a modest accuracy and forgetting gap to Progressive on the MNIST families, which closes as the support set grows.

\paragraph{Random-per-batch-mask ablations.}
Holding the FTN training pipeline fixed and replacing the support-conditioned configurer with a fresh random $k$-of-$H$ mask each step splits the configurer's role along input-shift type. Under correlated concept shift on the synthetic family the configurer is necessary even for retention, because the bank's atoms see colliding gradients across tasks that share latent components and only the configurer routes around the collisions. Under domain shift (Permuted MNIST, Rotated MNIST) random masks match FTN on retention, so the distributed bank with replay handles input-shift retention without the configurer; the configurer's contribution there is confined to head-start. On shared-shuffle MNIST (concept shift via component-shared label permutations) retention is nearly identical between FTN and the random-mask ablation while head-start is far worse without the configurer, because the bank absorbs the digit-label conflicts but only the configurer recovers a useful pre-update mask on a new permutation. 

\paragraph{Neither replay nor masking are sufficient for general retention.}
Removing the mask-overlap-filtered buffer markedly increases forgetting in the fixed-schedule reconfiguration models across every task family, with the largest jumps on the families whose tasks share latent structure: the mask alone is partial protection. Conversely, replay alone (NoMask\,+\,ER, matched memory) is sufficient under domain shift but collapses on shared-shuffle MNIST, since concept shift presents contradictory replay targets on the same input. FTN's combination of mask-filtered replay and per-batch reconfiguration is what produces strong retention and non-trivial transfer across all four families. High mean off-diagonal mask overlap occurs in both correlated and orthogonal synthetic regimes (\cref{app:mask_overlap}); the diagnostic signal is the correlation $r_{\mathrm{share}}$ between the model's pairwise-overlap matrix and the ground-truth concept-sharing matrix, not the absolute overlap level.


\paragraph{Limitations and future directions}
The largest limitation of the FTN models tested here is the extra computation required for reconfiguration of the router between gradient steps, particularly if used during inference.
In the FTN-MLP model, reconfiguration can take the fixed latent outputs of the MLPs and recompute only the final linear combination, and is thus cheap even for many router optimization steps.
In the FTN-LoRA model, the cost is much greater because a fixed, deep network serves as the combination function, and thus the entire model has to be evaluated at each step of router optimization.

One cheap alternative is to use the changepoint-driven replay-free versions, which we found here in the 16-task preliminary result and considered more extensively elsewhere\Cref{mckee2026cortex} to perform only slightly worse overall.
In theory, the drawbacks of the that version are (1) the lack of ensembling per-task, which is expected to both limit forward knowledge transfer and reduce robustness of recall to overwriting and search hardness, particularly in the LoRA model where the fixed network may have many layers or recurrent steps, and (2) no recourse to a replay buffer to repair regions of overlap among routing masks, further exacerbating the possibility of overwriting.
In practice, neither drawback presented strongly in our synthetic experiment.

Another way to improve the scalability of FTN is to reformulate it around approximate Bayesian updating of the mask.
That is, treat the mask logits as a distributional prior such that each reconfiguration stage need only produce a posterior update.
Changepoint detection can then be implemented as a conditionally flat prior.
At time of writing, we have implemented such an approach for large-scale problems and will present those results in a subsequent report.  
For the current study, we chose to focus on the theoretical grounding and present the original optimization-driven prototype for its algorithic simplicity.


\bibliography{refs}

@inproceedings{rahimi2007random,
  title={Random Features for Large-Scale Kernel Machines},
  author={Rahimi, Ali and Recht, Benjamin},
  booktitle={Advances in Neural Information Processing Systems},
  volume={20},
  year={2007},
  url={https://papers.nips.cc/paper_files/paper/2007/hash/013a006f03dbc5392effeb8f18fda755-Abstract.html}
}

@techreport{jaeger2001echo,
  title={The ``Echo State'' Approach to Analysing and Training Recurrent Neural Networks},
  author={Jaeger, Herbert},
  institution={German National Research Center for Information Technology},
  type={GMD Report},
  number={148},
  year={2001},
  url={https://publica-rest.fraunhofer.de/server/api/core/bitstreams/d789440a-da1a-4749-b225-49fd4035073a/content}
}

@article{lukosevicius2009reservoir,
  title={Reservoir Computing Approaches to Recurrent Neural Network Training},
  author={Luko{\v{s}}evi{\v{c}}ius, Mantas and Jaeger, Herbert},
  journal={Computer Science Review},
  volume={3},
  number={3},
  pages={127--149},
  year={2009},
  doi={10.1016/j.cosrev.2009.03.005}
}

@article{mccloskey1989catastrophic,
  title={Catastrophic interference in connectionist networks: The sequential learning problem},
  author={McCloskey, Michael and Cohen, Neal J},
  journal={Psychology of learning and motivation},
  volume={24},
  pages={109--165},
  year={1989},
  publisher={Elsevier}
}

@article{french1999catastrophic,
  title={Catastrophic forgetting in connectionist networks},
  author={French, Robert M},
  journal={Trends in cognitive sciences},
  volume={3},
  number={4},
  pages={128--135},
  year={1999},
  publisher={Elsevier}
}

@article{kirkpatrick2017overcoming,
  title={Overcoming catastrophic forgetting in neural networks},
  author={Kirkpatrick, James and Pascanu, Razvan and Rabinowitz, Neil and
          Veness, Joel and Desjardins, Guillaume and Rusu, Andrei A and
          Milan, Kieran and Quan, John and Ramalho, Tiago and Grabska-Barwinska,
          Agnieszka and others},
  journal={Proceedings of the national academy of sciences},
  volume={114},
  number={13},
  pages={3521--3526},
  year={2017},
  publisher={National Acad Sciences}
}

@inproceedings{zenke2017continual,
  title={Continual learning through synaptic intelligence},
  author={Zenke, Friedemann and Poole, Ben and Ganguli, Surya},
  booktitle={International Conference on Machine Learning},
  pages={3987--3995},
  year={2017},
  publisher={PMLR}
}

@article{lopez2017gradient,
  title={Gradient episodic memory for continual learning},
  author={Lopez-Paz, David and Ranzato, Marc'Aurelio},
  journal={Advances in neural information processing systems},
  volume={30},
  year={2017}
}

@inproceedings{rebuffi2017icarl,
  title={iCaRL: Incremental classifier and representation learning},
  author={Rebuffi, Sylvestre-Alvise and Kolesnikov, Alexander and Sperl, Georg
          and Lampert, Christoph H},
  booktitle={IEEE Conference on Computer Vision and Pattern Recognition (CVPR)},
  year={2017}
}

@article{wortsman2020supermasks,
  title={Supermasks in superposition},
  author={Wortsman, Mitchell and Ramanujan, Vivek and Liu, Rosanne and
          Kembhavi, Aniruddha and Rastegari, Mohammad and Yosinski, Jason
          and Farhadi, Ali},
  journal={Advances in Neural Information Processing Systems},
  volume={33},
  pages={15173--15184},
  year={2020}
}

@inproceedings{serra2018overcoming,
  title={Overcoming catastrophic forgetting with hard attention to the task},
  author={Serra, Joan and Suris, Didac and Miron, Marius and Karatzoglou, Alexandros},
  booktitle={International Conference on Machine Learning},
  pages={4548--4557},
  year={2018},
  organization={PMLR}
}

@inproceedings{mallya2018packnet,
  title={PackNet: Adding multiple tasks to a single network by iterative pruning},
  author={Mallya, Arun and Lazebnik, Svetlana},
  booktitle={IEEE Conference on Computer Vision and Pattern Recognition (CVPR)},
  year={2018}
}

@article{rusu2016progressive,
  title={Progressive neural networks},
  author={Rusu, Andrei A and Rabinowitz, Neil C and Desjardins, Guillaume and
          Soyer, Hubert and Kirkpatrick, James and Kavukcuoglu, Koray and
          Pascanu, Razvan and Hadsell, Raia},
  journal={arXiv preprint arXiv:1606.04671},
  year={2016}
}

@article{yang2019task,
  title={Task representations in neural networks trained to perform many
         different tasks},
  author={Yang, Guangyu Robert and Joglekar, Madhura R and Song, H Francis
          and Newsome, William T and Wang, Xiao-Jing},
  journal={Nature neuroscience},
  volume={22},
  number={2},
  pages={297--306},
  year={2019},
  publisher={Nature Publishing Group}
}

@article{mongillo2018inhibitory,
  title={Inhibitory connectivity defines the realm of excitatory plasticity},
  author={Mongillo, Gianluigi and Rumpel, Simon and Loewenstein, Yonatan},
  journal={Nature neuroscience},
  volume={21},
  number={10},
  pages={1463--1470},
  year={2018},
  publisher={Nature Publishing Group}
}

@article{cichon2015branch,
  title={Branch-specific dendritic Ca2+ spikes cause persistent synaptic
         plasticity},
  author={Cichon, Joseph and Gan, Wen-Biao},
  journal={Nature},
  volume={520},
  number={7546},
  pages={180--185},
  year={2015},
  publisher={Nature Publishing Group}
}

@article{poirazi2020illuminating,
  title={Illuminating dendritic function with computational models},
  author={Poirazi, Panayiota and Papoutsi, Athanasia},
  journal={Nature Reviews Neuroscience},
  volume={21},
  number={6},
  pages={303--321},
  year={2020},
  publisher={Nature Publishing Group}
}

@article{beniaguev2021single,
  title={Single cortical neurons as deep artificial neural networks},
  author={Beniaguev, David and Segev, Idan and London, Michael},
  journal={Neuron},
  volume={109},
  number={17},
  pages={2727--2739},
  year={2021},
  publisher={Elsevier}
}

@article{maass2000computational,
  title={On the computational power of winner-take-all},
  author={Maass, Wolfgang},
  journal={Neural computation},
  volume={12},
  number={11},
  pages={2519--2535},
  year={2000},
  publisher={MIT Press}
}

@article{douglas2004neuronal,
  title={Neuronal circuits of the neocortex},
  author={Douglas, Rodney J and Martin, Kevan AC},
  journal={Annual Review of Neuroscience},
  volume={27},
  pages={419--451},
  year={2004}
}

@article{rolls2010attractor,
  title={Attractor networks},
  author={Rolls, Edmund T},
  journal={Wiley Interdisciplinary Reviews: Cognitive Science},
  volume={1},
  number={1},
  pages={119--134},
  year={2010}
}

@article{shazeer2017outrageously,
  title={Outrageously large neural networks: The sparsely-gated mixture-of-experts layer},
  author={Shazeer, Noam and Mirhoseini, Azalia and Maziarz, Krzysztof and Davis, Andy and Le, Quoc and Hinton, Geoffrey and Dean, Jeff},
  journal={ICLR},
  year={2017}
}

@article{fedus2022switch,
  title={Switch transformers: Scaling to trillion parameter models with simple and efficient sparsity},
  author={Fedus, William and Zoph, Barret and Shazeer, Noam},
  journal={Journal of Machine Learning Research},
  volume={23},
  pages={1--39},
  year={2022}
}

@inproceedings{maddison2017concrete,
  title={The Concrete Distribution: A Continuous Relaxation of Discrete Random Variables},
  author={Maddison, Chris J. and Mnih, Andriy and Teh, Yee Whye},
  booktitle={International Conference on Learning Representations (ICLR)},
  year={2017}
}

@inproceedings{jang2017categorical,
  title={Categorical Reparameterization with Gumbel-Softmax},
  author={Jang, Eric and Gu, Shixiang and Poole, Ben},
  booktitle={International Conference on Learning Representations (ICLR)},
  year={2017}
}

@article{bengio2013estimating,
  title={Estimating or Propagating Gradients Through Stochastic Neurons for Conditional Computation},
  author={Bengio, Yoshua and L{\'e}onard, Nicholas and Courville, Aaron},
  journal={arXiv preprint arXiv:1308.3432},
  year={2013}
}

@inproceedings{chaudhry2018efficient,
  title={Efficient Lifelong Learning with A-{GEM}},
  author={Chaudhry, Arslan and Ranzato, Marc'Aurelio and Rohrbach, Marcus and Elhoseiny, Mohamed},
  booktitle={International Conference on Learning Representations (ICLR)},
  year={2019}
}

@inproceedings{abati2020conditional,
  title={Conditional Channel Gated Networks for Task-Aware Continual Learning},
  author={Abati, Davide and Tomczak, Jakub and Blankevoort, Tijmen and Calderara, Simone and Cucchiara, Rita and Bejnordi, Babak Ehteshami},
  booktitle={IEEE/CVF Conference on Computer Vision and Pattern Recognition (CVPR)},
  year={2020}
}

@article{poirazi2003pyramidal,
  title={Pyramidal Neuron as Two-Layer Neural Network},
  author={Poirazi, Panayiota and Brannon, Terrence and Mel, Bartlett W.},
  journal={Neuron},
  volume={37},
  number={6},
  pages={989--999},
  year={2003}
}

@article{amari1977dynamics,
  title={Dynamics of pattern formation in lateral-inhibition type neural fields},
  author={Amari, Shun-ichi},
  journal={Biological Cybernetics},
  volume={27},
  number={2},
  pages={77--87},
  year={1977}
}

@article{khona2022attractor,
  title={Attractor and integrator networks in the brain},
  author={Khona, Mikail and Fiete, Ila R.},
  journal={Nature Reviews Neuroscience},
  volume={23},
  number={12},
  pages={744--766},
  year={2022},
  publisher={Nature Publishing Group}
}

@article{faisal2008noise,
  title={Noise in the nervous system},
  author={Faisal, A. Aldo and Selen, Luc P. J. and Wolpert, Daniel M.},
  journal={Nature Reviews Neuroscience},
  volume={9},
  number={4},
  pages={292--303},
  year={2008},
  publisher={Nature Publishing Group}
}

@inproceedings{wang2022learning,
  title={Learning to Prompt for Continual Learning},
  author={Wang, Zifeng and Zhang, Zizhao and Lee, Chen-Yu and Zhang, Han and Sun, Ruoxi and Ren, Xiaoqi and Su, Guolong and Perot, Vincent and Dy, Jennifer and Pfister, Tomas},
  booktitle={CVPR},
  year={2022}
}

@inproceedings{wang2022dualprompt,
  title={DualPrompt: Complementary Prompting for Rehearsal-free Continual Learning},
  author={Wang, Zifeng and Zhang, Zizhao and Ebrahimi, Sayna and Sun, Ruoxi and Zhang, Han and Lee, Chen-Yu and Ren, Xiaoqi and Su, Guolong and Perot, Vincent and Dy, Jennifer and Pfister, Tomas},
  booktitle={ECCV},
  year={2022}
}

@article{fernando2017pathnet,
  title={PathNet: Evolution Channels Gradient Descent in Super Neural Networks},
  author={Fernando, Chrisantha and Banarse, Dylan and Blundell, Charles and Zwols, Yori and Ha, David and Rusu, Andrei A. and Pritzel, Alexander and Wierstra, Daan},
  journal={arXiv preprint arXiv:1701.08734},
  year={2017}
}

@inproceedings{veniat2021mntdp,
  title={Efficient Continual Learning with Modular Networks and Task-Driven Priors},
  author={Veniat, Tom and Denoyer, Ludovic and Ranzato, Marc'Aurelio},
  booktitle={ICLR},
  year={2021}
}

@inproceedings{javed2019oml,
  title={Meta-Learning Representations for Continual Learning},
  author={Javed, Khurram and White, Martha},
  booktitle={NeurIPS},
  year={2019}
}

@inproceedings{buzzega2020dark,
  title={Dark Experience for General Continual Learning: A Strong, Simple Baseline},
  author={Buzzega, Pietro and Boschini, Matteo and Porrello, Angelo and Abati, Davide and Calderara, Simone},
  booktitle={NeurIPS},
  year={2020}
}

@article{georgopoulos1986neuronal,
  title={Neuronal Population Coding of Movement Direction},
  author={Georgopoulos, Apostolos P. and Schwartz, Andrew B. and Kettner, Ronald E.},
  journal={Science},
  volume={233},
  number={4771},
  pages={1416--1419},
  year={1986},
  doi={10.1126/science.3749885},
  publisher={American Association for the Advancement of Science}
}

@article{averbeck2006neural,
  title={Neural Correlations, Population Coding and Computation},
  author={Averbeck, Bruno B. and Latham, Peter E. and Pouget, Alexandre},
  journal={Nature Reviews Neuroscience},
  volume={7},
  number={5},
  pages={358--366},
  year={2006},
  doi={10.1038/nrn1888},
  publisher={Nature Publishing Group}
}

@article{cohen2011measuring,
  title={Measuring and Interpreting Neuronal Correlations},
  author={Cohen, Marlene R. and Kohn, Adam},
  journal={Nature Neuroscience},
  volume={14},
  number={7},
  pages={811--819},
  year={2011},
  doi={10.1038/nn.2842},
  publisher={Nature Publishing Group}
}

@article{tolhurst1983statistical,
  title={The Statistical Reliability of Signals in Single Neurons in Cat and Monkey Visual Cortex},
  author={Tolhurst, D. J. and Movshon, J. A. and Dean, A. F.},
  journal={Vision Research},
  volume={23},
  number={8},
  pages={775--785},
  year={1983},
  doi={10.1016/0042-6989(83)90200-6},
  publisher={Elsevier}
}

@article{softky1993highly,
  title={The Highly Irregular Firing of Cortical Cells Is Inconsistent with Temporal Integration of Random {EPSPs}},
  author={Softky, William R. and Koch, Christof},
  journal={Journal of Neuroscience},
  volume={13},
  number={1},
  pages={334--350},
  year={1993},
  doi={10.1523/JNEUROSCI.13-01-00334.1993}
}

@article{shadlen1998variable,
  title={The Variable Discharge of Cortical Neurons: Implications for Connectivity, Computation, and Information Coding},
  author={Shadlen, Michael N. and Newsome, William T.},
  journal={Journal of Neuroscience},
  volume={18},
  number={10},
  pages={3870--3896},
  year={1998},
  doi={10.1523/JNEUROSCI.18-10-03870.1998}
}

@article{churchland2010stimulus,
  title={Stimulus Onset Quenches Neural Variability: A Widespread Cortical Phenomenon},
  author={Churchland, Mark M. and Yu, Byron M. and Cunningham, John P. and Sugrue, Leo P. and Cohen, Marlene R. and Corrado, Greg S. and Newsome, William T. and Clark, Andrew M. and Hosseini, Paymon and Scott, Benjamin B. and Bradley, David C. and Smith, Matthew A. and Kohn, Adam and Movshon, J. Anthony and Armstrong, Katherine M. and Moore, Tirin and Chang, Steve W. and Snyder, Lawrence H. and Lisberger, Stephen G. and Priebe, Nicholas J. and Finn, Ian M. and Ferster, David and Ryu, Stephen I. and Santhanam, Gopal and Sahani, Maneesh and Shenoy, Krishna V.},
  journal={Nature Neuroscience},
  volume={13},
  number={3},
  pages={369--378},
  year={2010},
  doi={10.1038/nn.2501},
  publisher={Nature Publishing Group}
}

@article{goris2014partitioning,
  title={Partitioning Neuronal Variability},
  author={Goris, Robbe L. T. and Movshon, J. Anthony and Simoncelli, Eero P.},
  journal={Nature Neuroscience},
  volume={17},
  number={6},
  pages={858--865},
  year={2014},
  doi={10.1038/nn.3711},
  publisher={Nature Publishing Group}
}

@article{lecun1998mnist,
  title={Gradient-based learning applied to document recognition},
  author={LeCun, Yann and Bottou, L{\'e}on and Bengio, Yoshua and Haffner, Patrick},
  journal={Proceedings of the IEEE},
  volume={86},
  number={11},
  pages={2278--2324},
  year={1998}
}

@article{mckee2026cortex,
  title={Cortex-Inspired Continual Learning: Unsupervised Instantiation and Recovery of Functional Task Networks},
  author={McKee, Kevin and Hazy, Thomas and Zheng, Yicong and Bugaud, Zacharie and Miconi, Thomas},
  journal={arXiv preprint arXiv:2604.24637},
  year={2026}
}
\bibliographystyle{plainnat}

\newpage
\appendix

\section{Implementation details}\label{app:impl}

\paragraph{Configurers.}
All three configurers expose the same external interface and differ in the soft-mask form, optimizer, logit initialization, and auxiliary loss. AnnealSTE uses SGD, zero initialization, hard $k$-WTA with a straight-through estimator, and no auxiliary term. Sigmoid uses RMSprop, zero initialization, a soft mask $\sigma(m)$, and post-hoc top-$k$ projection. Gumbel-Sigmoid uses Adam, initialization at $\sigma^{-1}(k/H)$, $\sigma((m+g_i)/\tau_i)$ with logistic noise, and the budget penalty $(\sum\sigma(m)-k)^2/k^2$.

\paragraph{Replay buffer.}
The replay buffer stores 200 $(x,y,m)$ triples. On every gradient step we sample one triple whose stored mask has fractional overlap $\rho=(m_\mathrm{cur}\cdot m_j)/\|m_\mathrm{cur}\|_1\in(0.2,0.8]$ with the current mask. Entries with $\rho\le0.2$ are skipped because they are unlikely to share useful computation with the current update; entries with $\rho>0.8$ are skipped because the current minibatch already trains a very similar mask.

\paragraph{Reconfig protocol.}
At evaluation time we draw a labeled support set from the target task and run the configurer on it to obtain a recall mask. The model is not told which task generated the support set. Metrics are computed on a larger held-out batch from the same task, disjoint from the support set.

\paragraph{SupSup baseline.}
We instantiate SupSup~\citep{wortsman2020supermasks} on the same parallel-MLP bank ($H=8192$ slots, per-MLP topology identical to FTN's). The bank's parallel-MLP weights are initialized once and \emph{frozen} for the rest of training; dropout on the bank output is disabled for SupSup so the frozen features are deterministic. For each task $t$ we instantiate (i) a fresh score vector $s_t\in\mathbb{R}^H$ and (ii) a fresh per-task readout head $W_t \in \mathbb{R}^{\mathrm{out}\times H}$ initialized by Xavier-uniform. The forward pass uses a hard top-$k$ mask via the straight-through estimator $\mathrm{mask} = \mathrm{topk}(s_t, k) + s_t - \mathrm{detach}(s_t)$; both $s_t$ and $W_t$ receive gradient through the task loss, the bank does not. Optimisers: Adam, $\mathrm{lr}=10^{-2}$ on $s_t$ and $3\times10^{-4}$ on $W_t$. Per-task storage is the pair $(\mathrm{topk}(s_t,k), W_t)$. At evaluation (no task identity) the support-set retrieval scans every stored pair, swaps in the candidate readout, computes the support cross-entropy, and returns the lowest-loss pair; the held-out batch is then evaluated under that pair. This is the deterministic, one-hot version of SupSup's GG retrieval.

\paragraph{Progressive Networks baseline.}
We instantiate Progressive Networks~\citep{rusu2016progressive} on the same parallel-MLP bank. Task $t$ owns a disjoint $k$-block ``column'' covering slots $[t\cdot k,\,(t+1)\cdot k)$. During training of task $t$: (i) the parallel-MLP weights for slots outside that range are \emph{frozen} (gradients zeroed before the optimizer step), so all prior columns are untouched; (ii) the active mask is the cumulative mask $m_t = \mathbb{1}\!\left[\,\mathrm{slot} < (t+1)\cdot k\,\right]$, so the per-task readout head $W_t$ takes input from the new column \emph{and} all previously trained frozen columns (the lateral channel); (iii) only $W_t$ and the new-column slot weights receive gradient. Per-task storage is the pair $(m_t, W_t)$ where $m_t$ is the cumulative mask. At evaluation the same support-set retrieval as SupSup is used: scan stored pairs, return the lowest-support-loss pair, evaluate held-out under it. Hyperparameters match FTN ($k=512$, Adam $\mathrm{lr}=3\times10^{-4}$, no replay). For task counts up to $\lfloor H/k \rfloor = 16$ no slot wraps; the four benchmarks here use 6, 10, 10, and 12 tasks respectively, so column allocation never collides. This adaptation preserves the two defining properties of Progressive Networks (frozen prior capacity and lateral access for forward transfer); the only deviations from \citet{rusu2016progressive} are that the columns are sub-blocks of a single backbone (rather than separately allocated networks) and that we use support-set retrieval at evaluation rather than task identity, since our protocol forbids task identity.

\section{Per-task performance matrix snapshots}\label{app:perf}
We include the lower-triangular performance matrices $R$ at the default configuration on both task families.

\begin{figure}[ht]
\centering
\begin{subfigure}[b]{0.48\columnwidth}
\includegraphics[width=\textwidth]{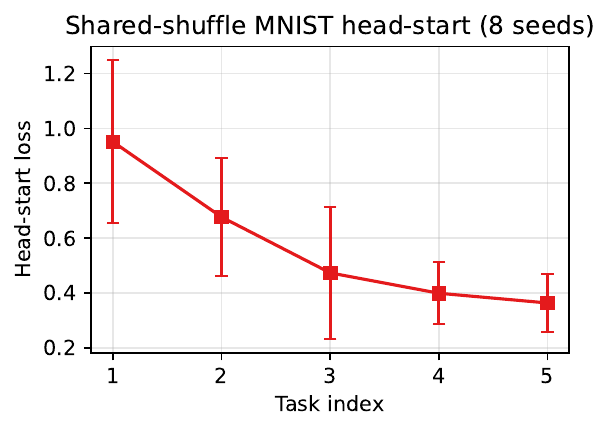}
\caption{Head-start per task, $S=30$.}
\end{subfigure}\hfill
\begin{subfigure}[b]{0.48\columnwidth}
\includegraphics[width=\textwidth]{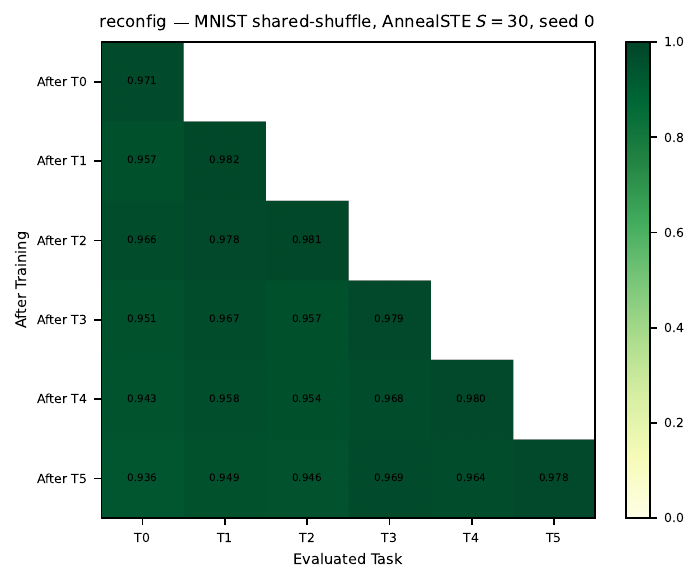}
\caption{Reconfig protocol performance matrix.}
\end{subfigure}
\caption{Shared-shuffle MNIST with the default FTN configuration over 8 seeds. The top-$k$ mask supports continual learning under concept shift; the reconfig protocol recovers prior tasks from a small support set without task identity.}
\label{fig:mnist}
\end{figure}

\section{Secondary Experiments}\label{app:sec_exp}

\subsection{Primary search and baselines}\label{sec:exp1}
\begin{figure}[t]
\centering
\begin{subfigure}[b]{0.32\textwidth}
\includegraphics[width=\textwidth]{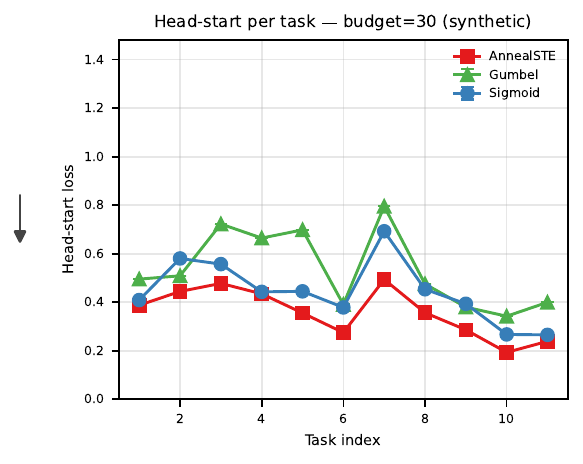}
\caption{Head-start per task, $S=30$.}
\end{subfigure}\hfill
\begin{subfigure}[b]{0.32\textwidth}
\includegraphics[width=\textwidth]{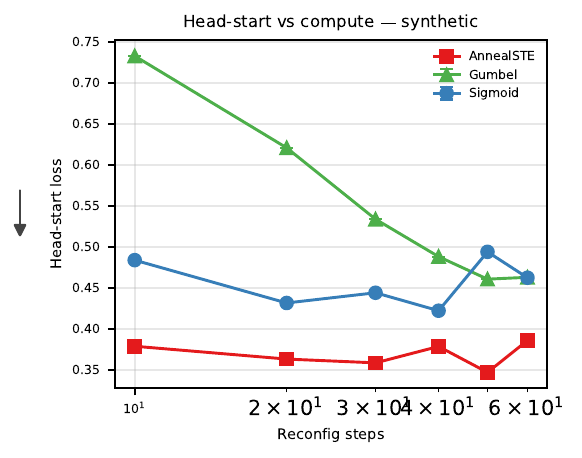}
\caption{Head-start vs. budget.}
\end{subfigure}\hfill
\begin{subfigure}[b]{0.32\textwidth}
\includegraphics[width=\textwidth]{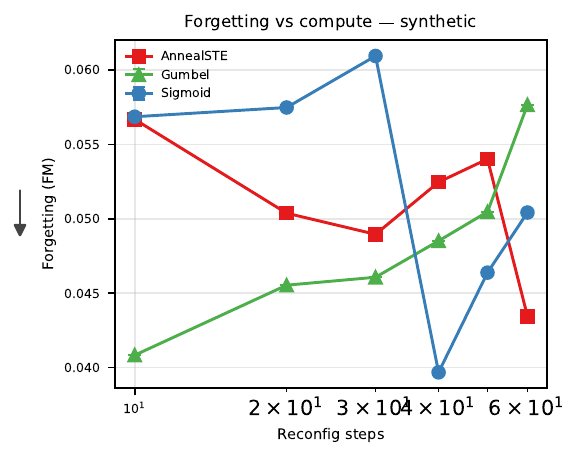}
\caption{Forgetting vs. budget.}
\end{subfigure}
\caption{Head-start loss and forgetting on the correlated synthetic family. Lower is better in both metrics. AnnealSTE obtains strong head-start loss at small budget, and all configurers reach low forgetting under per-batch reconfiguration.}
\label{fig:headstart_compute}
\end{figure}

We first sweep configurer type in \{Sigmoid, AnnealSTE, Gumbel-Sigmoid\} and inner-step budget $S\in\{10,20,30,40,50,60\}$ on the correlated synthetic family with $k=512$, support size $256$, and per-batch reconfiguration. The sweep uses one shared seed to compare the 18 configurations within seed; subsequent ablations report mean and standard deviation over 8 seeds. All three configurers reach low forgetting (FM in $[0.04,0.06]$ across the sweep) and similar accuracy (ACC in $[0.89,0.92]$). The discriminating axis is head-start: AnnealSTE attains the best head-start loss at every budget (mean $0.37$ across $S$, vs.\ $0.46$ for Sigmoid and $0.55$ for Gumbel-Sigmoid) and is stable across budgets, while Gumbel-Sigmoid improves steadily with $S$ but does not reach AnnealSTE's level. AnnealSTE at $S=30$ gives head-start $0.358$, $\mathrm{FM}=0.049$, $r_{\mathrm{share}}=0.509$; doubling the budget to $S=60$ yields no head-start gain ($0.386$) at twice the inner-loop cost. We therefore adopt AnnealSTE, $S=30$, $k=512$, support size $256$, and per-batch reconfiguration as the default. The full $3\times 6$ sweep table is in \cref{app:primary_sweep}; \cref{fig:headstart_compute} plots head-start and forgetting against budget.

\Cref{tab:baselines_synthetic,tab:baselines_mnist} compare FTN with two endpoints using the same parallel bank. NoMask trains the full bank on every task, representing full sharing. FixedMask assigns deterministic disjoint blocks of $k$ neurons, representing oracle isolation with no mask inference or sharing. FTN lies between these endpoints. On synthetic tasks, NoMask has $\mathrm{FM}=0.392\pm0.016$, FixedMask has $0$ by construction, and FTN obtains $0.055\pm0.015$, closing about $86\%$ of the forgetting gap. On shared-shuffle MNIST, FTN closes about $97\%$ of the gap. Unlike FixedMask, FTN also improves head-start loss on correlated tasks and recovers nonzero concept-sharing structure.

\begin{table}[t]
\centering
\footnotesize
\caption{Baseline comparison on synthetic. NoMask trains the full network without any per-task mask (full parameter sharing); FixedMask uses deterministic disjoint blocks of size $k$ per task (no transfer, oracle no-forgetting). FTN closes the FM gap between NoMask and FixedMask while recovering non-trivial concept-sharing structure that both baselines lack ($r_{\mathrm{share}}=0$ by construction). Mean$\pm$std over 8 seeds.}
\label{tab:baselines_synthetic}
\begin{tabular}{lccccc}
\toprule
Method & ACC$\uparrow$ & FM$\downarrow$ & BWT$\uparrow$ & Head-start$\downarrow$ & $r_{\mathrm{share}}\uparrow$ \\
\midrule
NoMask & 0.610\,$\pm$\,0.019 & 0.392\,$\pm$\,0.016 & -0.392\,$\pm$\,0.016 & 2.830\,$\pm$\,0.293 & 0.000\,$\pm$\,0.000 \\
FixedMask & 0.945\,$\pm$\,0.020 & 0.000\,$\pm$\,0.000 & +0.000\,$\pm$\,0.000 & 0.702\,$\pm$\,0.003 & 0.000\,$\pm$\,0.000 \\
FTN (AnnealSTE, $S{=}30$) & 0.886\,$\pm$\,0.042 & 0.055\,$\pm$\,0.015 & -0.053\,$\pm$\,0.017 & 0.423\,$\pm$\,0.077 & 0.454\,$\pm$\,0.213 \\
\bottomrule
\end{tabular}
\end{table}

\begin{table}[t]
\centering
\footnotesize
\caption{Baseline comparison on MNIST shared-shuffle. NoMask trains the full network without any per-task mask (full parameter sharing); FixedMask uses deterministic disjoint blocks of size $k$ per task (no transfer, oracle no-forgetting). FTN closes the FM gap between NoMask and FixedMask while recovering non-trivial concept-sharing structure that both baselines lack ($r_{\mathrm{share}}=0$ by construction). Mean$\pm$std over 8 seeds.}
\label{tab:baselines_mnist}
\begin{tabular}{lccccc}
\toprule
Method & ACC$\uparrow$ & FM$\downarrow$ & BWT$\uparrow$ & Head-start$\downarrow$ & $r_{\mathrm{share}}\uparrow$ \\
\midrule
NoMask & 0.433\,$\pm$\,0.056 & 0.656\,$\pm$\,0.067 & -0.656\,$\pm$\,0.067 & 10.840\,$\pm$\,0.881 & 0.000\,$\pm$\,0.000 \\
FixedMask & 0.978\,$\pm$\,0.001 & 0.000\,$\pm$\,0.000 & +0.000\,$\pm$\,0.000 & 2.312\,$\pm$\,0.016 & 0.000\,$\pm$\,0.000 \\
FTN (AnnealSTE, $S{=}30$) & 0.962\,$\pm$\,0.004 & 0.021\,$\pm$\,0.005 & -0.021\,$\pm$\,0.005 & 0.573\,$\pm$\,0.039 & 0.398\,$\pm$\,0.127 \\
\bottomrule
\end{tabular}
\end{table}

\subsection{Reconfiguration frequency}\label{sec:exp_freq}

Per-batch reconfiguration is central to FTN's transfer behavior. A natural alternative is to configure once at the beginning of a task block and reuse that mask for the entire task. This per-task variant creates cleaner stored masks, but it also trains a narrower set of subnetworks. In contrast, per-batch reconfiguration exposes many nearby masks to gradient signal, creating a broader pool of partially trained assemblies that future support-set optimizations can recombine.

\begin{figure}[t]
\centering
\begin{subfigure}[b]{0.9\textwidth}
\includegraphics[width=\textwidth]{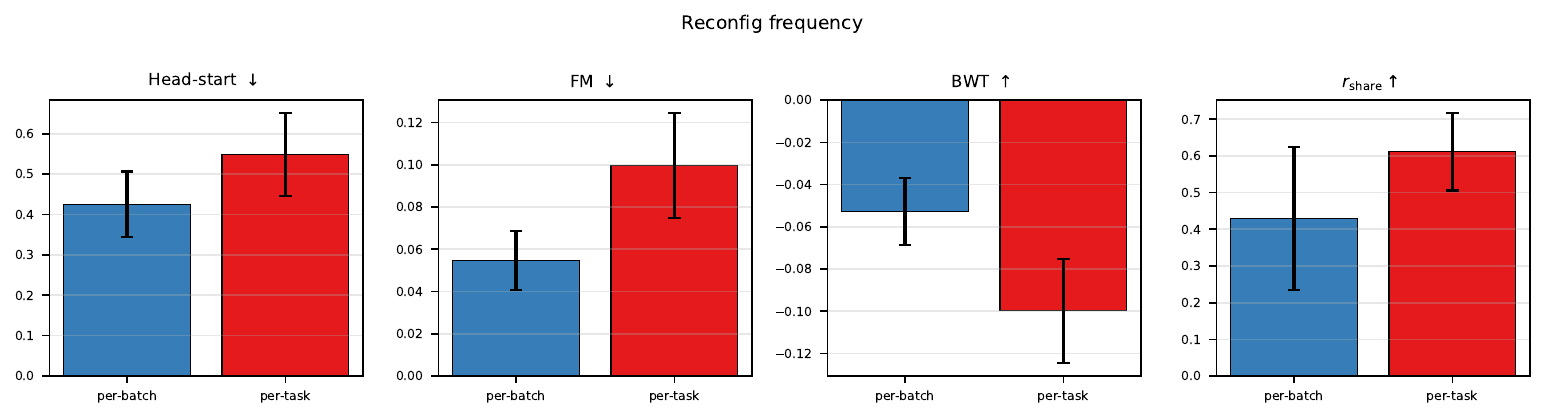}
\caption{Head-start, FM, BWT, and sharing fidelity.}
\end{subfigure}

\begin{subfigure}[b]{0.6\textwidth}
\includegraphics[width=\textwidth]{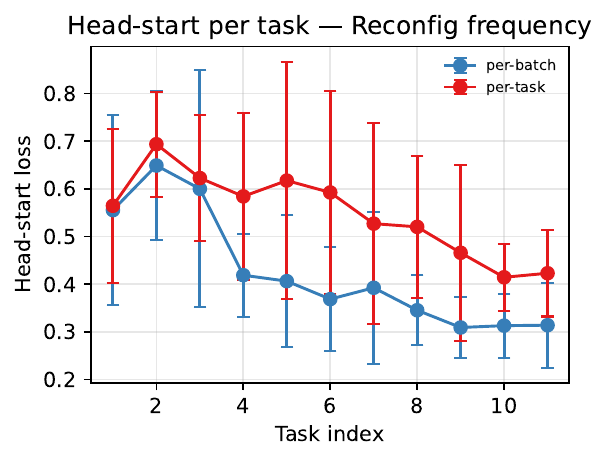}
\caption{Head-start vs. task index.}
\end{subfigure}
\caption{Reconfiguration frequency on the correlated synthetic family over 8 seeds. Per-batch reconfiguration lowers head-start loss relative to one mask per task, supporting the view that mask variation during training builds a broader reusable basis.}
\label{fig:exp3_freq}
\end{figure}

Empirically, per-batch reconfiguration gives lower head-start loss and lower forgetting on the correlated synthetic family (\cref{fig:exp3_freq}). Per-batch obtains head-start loss $0.425\pm0.081$ and $\mathrm{FM}=0.055\pm0.014$, while per-task gives $0.548\pm0.103$ and $0.100\pm0.025$. Per-task has higher $r_{\mathrm{share}}$ ($0.611\pm0.105$ versus $0.429\pm0.195$), consistent with a more stable stored mask, but the per-batch regime is better for forward transfer under support-set reconfiguration.

\subsection{Sparsity and support-set size}\label{sec:exp_k_b}
\begin{figure}[t]
\centering
\begin{subfigure}[b]{0.9\textwidth}
\includegraphics[width=\textwidth]{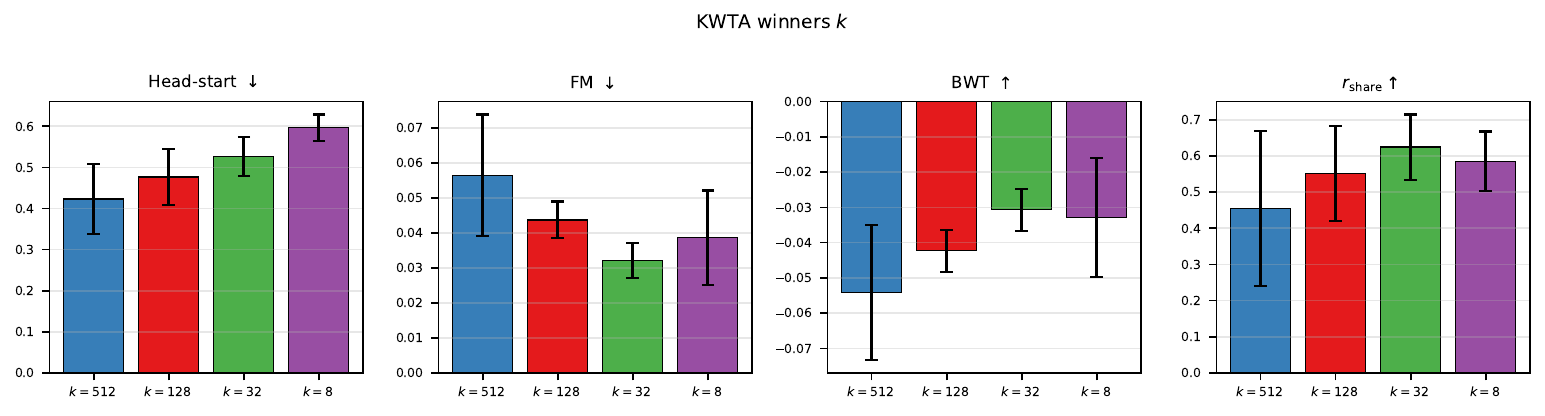}
\caption{Sweep over $k$.}
\end{subfigure}

\begin{subfigure}[b]{0.6\textwidth}
\includegraphics[width=\textwidth]{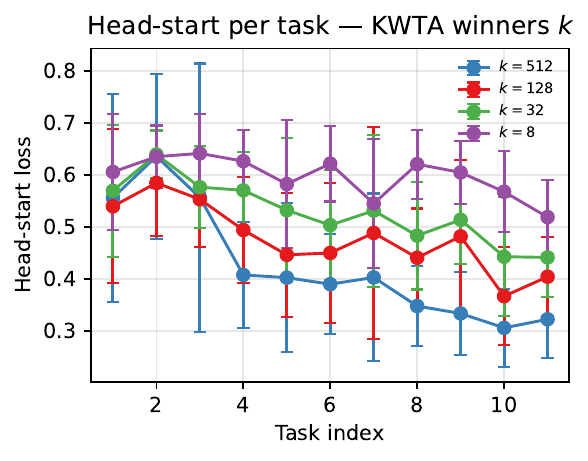}
\caption{Head-start vs.\ task index, by $k$.}
\end{subfigure}
\caption{Head-start loss, FM, BWT, and sharing fidelity versus top-$k$ budget $k$ on the correlated synthetic family over 8 seeds. The per-task head-start curve shows how forward transfer evolves as the bank fills out across tasks. Detailed tables are in \cref{app:ablation_tables}.}
\label{fig:exp4}
\end{figure}

The top-$k$ budget controls how much overlap is possible. Sweeping $k\in\{512,128,32,8\}$ shows a clear isolation-transfer trade-off (\cref{fig:exp4}). Reducing $k$ lowers FM from $0.056$ at $k=512$ to $0.032$ at $k=32$, then $0.039$ at $k=8$, while head-start loss rises from about $0.42$ to $0.60$. Sharing fidelity improves from $0.455$ to $0.624$ as $k$ decreases to $32$, then falls slightly at $k=8$. Moderate sparsity gives the cleanest structured overlap; extreme sparsity improves isolation but leaves less reusable basis for future tasks.

\begin{figure}[t]
\centering
\begin{subfigure}[b]{0.9\textwidth}
\includegraphics[width=\textwidth]{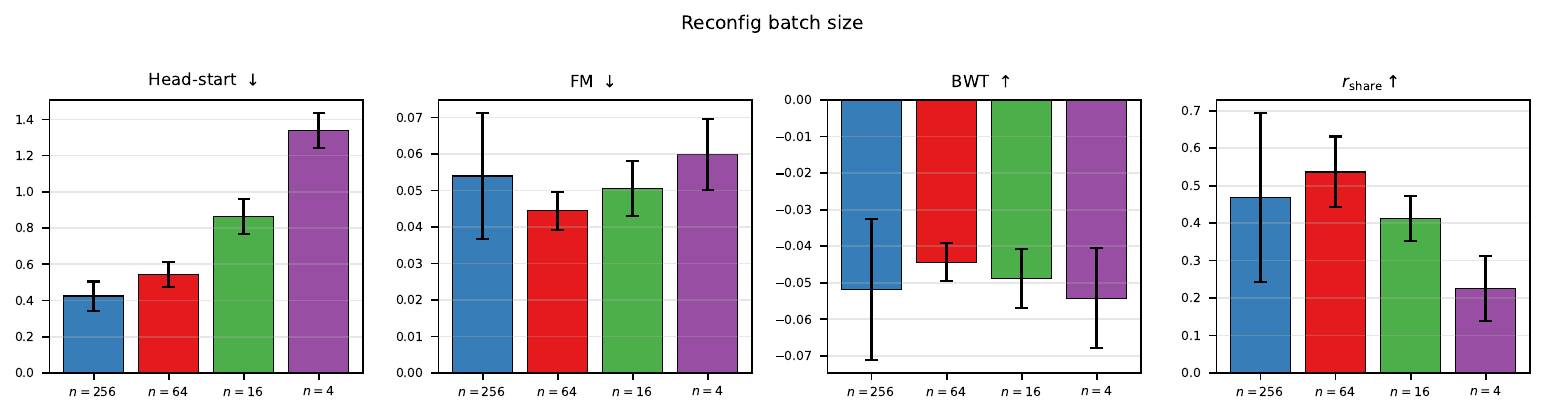}
\caption{Sweep over support size.}
\end{subfigure}

\begin{subfigure}[b]{0.6\textwidth}
\includegraphics[width=\textwidth]{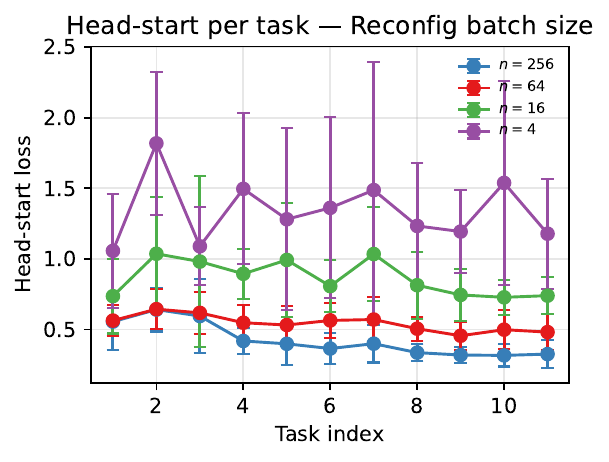}
\caption{Head-start vs.\ task index, by support size.}
\end{subfigure}
\caption{Head-start loss, FM, BWT, and sharing fidelity versus support-set size on the correlated synthetic family over 8 seeds. The per-task head-start curve shows how forward transfer evolves as the bank fills out across tasks. Detailed tables are in \cref{app:ablation_tables}.}
\label{fig:exp5}
\end{figure}

The support size used by the configurer affects transfer more than retention (\cref{fig:exp5}). When support size decreases from $256$ to $4$, head-start loss rises from $0.425\pm0.080$ to $1.339\pm0.095$, reflecting poorer generalization from the support set to held-out examples. In contrast, FM remains at or below $0.060$ across the sweep, and ACC stays in the $0.86$ to $0.89$ range. Thus the continual-learning protection is robust to small supports, while forward transfer depends on enough examples to infer a useful mask.

\section{Primary configurer sweep}\label{app:primary_sweep}
\Cref{tab:main_synthetic} reports the full $3\times 6$ sweep over configurer type and inner-step budget summarised in \cref{sec:exp1}. All three configurers reach low forgetting (FM\,$\in [0.04, 0.06]$) and comparable accuracy; the discriminating axis is head-start loss, where AnnealSTE attains the lowest values at every budget and Gumbel-Sigmoid improves with $S$ but does not catch up. AnnealSTE at $S=30$ is the chosen default: head-start $0.358$, $\mathrm{FM}=0.049$, $r_{\mathrm{share}}=0.509$, with no head-start gain from doubling the budget.
\begin{table}[t]
\caption{Knowledge-transfer results on \texttt{synthetic} (single shared seed across all 18 method, budget configurations): ACC, forgetting (FM), backward transfer (BWT), neuron-sharing fidelity ($r_{\mathrm{share}}$), and pre-training mask-only loss (head-start) averaged over tasks 2 to N.}
\label{tab:main_synthetic}
\centering
\small
\begin{tabular}{llccccc}
\toprule
Method & Budget & ACC$\uparrow$ & FM$\downarrow$ & BWT$\uparrow$ & Head-start$\downarrow$ & $r_{\mathrm{share}}\uparrow$\\
\midrule
Sigmoid & 10 & 0.900 & 0.057 & -0.057 & 0.484 & 0.491 \\
Sigmoid & 20 & 0.897 & 0.057 & -0.057 & 0.432 & 0.581 \\
Sigmoid & 30 & 0.894 & 0.061 & -0.061 & 0.444 & 0.477 \\
Sigmoid & 40 & 0.920 & 0.040 & -0.040 & 0.422 & 0.226 \\
Sigmoid & 50 & 0.914 & 0.046 & -0.046 & 0.494 & 0.221 \\
Sigmoid & 60 & 0.912 & 0.050 & -0.050 & 0.462 & 0.262 \\
AnnealSTE & 10 & 0.907 & 0.057 & -0.057 & 0.379 & 0.622 \\
AnnealSTE & 20 & 0.913 & 0.050 & -0.050 & 0.363 & 0.564 \\
AnnealSTE & 30 & 0.909 & 0.049 & -0.049 & 0.358 & 0.509 \\
AnnealSTE & 40 & 0.913 & 0.052 & -0.052 & 0.378 & 0.617 \\
AnnealSTE & 50 & 0.909 & 0.054 & -0.054 & 0.347 & 0.575 \\
AnnealSTE & 60 & 0.916 & 0.043 & -0.043 & 0.386 & 0.601 \\
Gumbel & 10 & 0.893 & 0.041 & -0.041 & 0.733 & 0.422 \\
Gumbel & 20 & 0.908 & 0.046 & -0.045 & 0.621 & 0.179 \\
Gumbel & 30 & 0.911 & 0.046 & -0.046 & 0.534 & 0.365 \\
Gumbel & 40 & 0.910 & 0.049 & -0.049 & 0.488 & 0.484 \\
Gumbel & 50 & 0.904 & 0.050 & -0.050 & 0.461 & 0.507 \\
Gumbel & 60 & 0.906 & 0.058 & -0.058 & 0.463 & 0.508 \\
\bottomrule
\end{tabular}
\end{table}

\section{Detailed ablation tables}\label{app:ablation_tables}
This appendix collects the numerical tables underlying the secondary-experiment plots in \cref{app:sec_exp}. The tables cover the task-structure ablation contrasting the correlated synthetic family with an orthogonal control (under orthogonal task generation the configurer's mask-sharing structure collapses to chance, whereas under correlated generation it tracks the ground-truth concept-sharing matrix), the per-batch versus per-task reconfiguration comparison on the correlated synthetic family (showing the head-start and forgetting penalty incurred by holding the mask fixed across a task block), and the top-$k$ budget and support-size sweeps that source the trade-off curves in \cref{fig:exp4,fig:exp5}. All entries are mean$\pm$std over 8 seeds at the default AnnealSTE configuration ($S{=}30$, $k{=}512$, support size $256$, except along the swept axis).

\section{Per-benchmark tables (full mean$\pm$std)}\label{app:revision_tables}
The body \cref{tab:mnist_suite} and the synthetic \cref{tab:revision_synthetic} compress the per-benchmark tables to means; the full mean$\pm$std tables, including FTN+Router and no-replay diagnostics for every family, are below. Each table covers one MNIST task family (shared-shuffle, Permuted, Rotated) and reports ACC, FM, BWT, and head-start NLL with standard deviations across seeds for every comparator. These are the source tables for the per-family claims in \cref{sec:results_mnist} and the discussion of how the configurer's contribution shifts between concept- and domain-shift regimes.
\begin{table}[t]
\centering
\footnotesize
\caption{Shared-shuffle MNIST (6 tasks) under the reconfig (no-task-identity) protocol. Mean$\pm$std over 5 seeds. Higher ACC and BWT, lower FM and head-start (HS) loss, are better.}
\label{tab:revision_mnist_shuffle}
\begin{tabular}{lllll}
\toprule
Method & ACC$\uparrow$ & FM$\downarrow$ & BWT$\uparrow$ & HS$\downarrow$ \\
\midrule
NoMask (naive seq.) & $0.444 \pm 0.065$ & $0.644 \pm 0.077$ & $-0.644 \pm 0.077$ & $10.397 \pm 0.811$ \\
NoMask + ER (random replay) & $0.442 \pm 0.064$ & $0.630 \pm 0.075$ & $-0.630 \pm 0.075$ & $8.524 \pm 0.625$ \\
FixedMask (oracle ID at eval) & $0.979 \pm 0.000$ & $0.000 \pm 0.000$ & $0.000 \pm 0.000$ & $2.309 \pm 0.018$ \\
FixedMask + retrieval router & $0.978 \pm 0.000$ & $0.000 \pm 0.000$ & $0.000 \pm 0.000$ & $2.311 \pm 0.017$ \\
FTN, random per-batch masks (ablation) & $0.923 \pm 0.006$ & $0.027 \pm 0.006$ & $-0.024 \pm 0.005$ & $6.652 \pm 0.628$ \\
FTN (ours) & $0.960 \pm 0.005$ & $0.023 \pm 0.005$ & $-0.023 \pm 0.005$ & $0.562 \pm 0.028$ \\
FTN + retrieval router (eval-only) & $0.961 \pm 0.002$ & $0.022 \pm 0.003$ & $-0.022 \pm 0.003$ & $0.561 \pm 0.019$ \\
FTN, no replay & $0.881 \pm 0.021$ & $0.119 \pm 0.025$ & $-0.119 \pm 0.025$ & $0.761 \pm 0.053$ \\
\bottomrule
\end{tabular}
\end{table}

\begin{table}[t]
\centering
\footnotesize
\caption{PermutedMNIST (10 tasks) under the reconfig (no-task-identity) protocol. Mean$\pm$std over 5 seeds. Higher ACC and BWT, lower FM and head-start (HS) loss, are better.}
\label{tab:revision_mnist_permuted}
\begin{tabular}{lllll}
\toprule
Method & ACC$\uparrow$ & FM$\downarrow$ & BWT$\uparrow$ & HS$\downarrow$ \\
\midrule
NoMask (naive seq.) & $0.928 \pm 0.007$ & $0.053 \pm 0.008$ & $-0.053 \pm 0.008$ & $5.207 \pm 0.195$ \\
NoMask + ER (random replay) & $0.960 \pm 0.002$ & $0.018 \pm 0.002$ & $-0.018 \pm 0.002$ & $5.270 \pm 0.153$ \\
FixedMask (oracle ID at eval) & $0.979 \pm 0.000$ & $0.000 \pm 0.000$ & $0.000 \pm 0.000$ & $2.321 \pm 0.005$ \\
FixedMask + retrieval router & $0.979 \pm 0.000$ & $0.000 \pm 0.000$ & $0.000 \pm 0.000$ & $2.320 \pm 0.006$ \\
FTN, random per-batch masks (ablation) & $0.915 \pm 0.003$ & $0.017 \pm 0.004$ & $-0.016 \pm 0.004$ & $3.411 \pm 0.096$ \\
FTN (ours) & $0.954 \pm 0.001$ & $0.024 \pm 0.001$ & $-0.024 \pm 0.001$ & $1.616 \pm 0.026$ \\
FTN + retrieval router (eval-only) & $0.952 \pm 0.004$ & $0.027 \pm 0.004$ & $-0.027 \pm 0.004$ & $1.593 \pm 0.013$ \\
FTN, no replay & $0.945 \pm 0.001$ & $0.034 \pm 0.001$ & $-0.034 \pm 0.001$ & $1.577 \pm 0.029$ \\
\bottomrule
\end{tabular}
\end{table}

\begin{table}[t]
\centering
\footnotesize
\caption{RotatedMNIST (10 tasks) under the reconfig (no-task-identity) protocol. Mean$\pm$std over 5 seeds. Higher ACC and BWT, lower FM and head-start (HS) loss, are better.}
\label{tab:revision_mnist_rotated}
\begin{tabular}{lllll}
\toprule
Method & ACC$\uparrow$ & FM$\downarrow$ & BWT$\uparrow$ & HS$\downarrow$ \\
\midrule
NoMask (naive seq.) & $0.523 \pm 0.007$ & $0.509 \pm 0.008$ & $-0.509 \pm 0.008$ & $0.360 \pm 0.010$ \\
NoMask + ER (random replay) & $0.912 \pm 0.003$ & $0.075 \pm 0.003$ & $-0.075 \pm 0.003$ & $0.667 \pm 0.025$ \\
FixedMask (oracle ID at eval) & $0.979 \pm 0.000$ & $0.000 \pm 0.000$ & $0.000 \pm 0.000$ & $2.322 \pm 0.007$ \\
FixedMask + retrieval router & $0.979 \pm 0.000$ & $0.000 \pm 0.000$ & $0.000 \pm 0.000$ & $2.321 \pm 0.007$ \\
FTN, random per-batch masks (ablation) & $0.867 \pm 0.002$ & $0.087 \pm 0.002$ & $-0.087 \pm 0.002$ & $0.621 \pm 0.030$ \\
FTN (ours) & $0.941 \pm 0.000$ & $0.043 \pm 0.001$ & $-0.043 \pm 0.001$ & $0.324 \pm 0.007$ \\
FTN + retrieval router (eval-only) & $0.936 \pm 0.003$ & $0.049 \pm 0.003$ & $-0.049 \pm 0.003$ & $0.311 \pm 0.002$ \\
FTN, no replay & $0.767 \pm 0.004$ & $0.238 \pm 0.004$ & $-0.238 \pm 0.004$ & $0.275 \pm 0.005$ \\
\bottomrule
\end{tabular}
\end{table}

\section{Mask overlap statistics}\label{app:mask_overlap}
\Cref{tab:mask_overlap} reports the pairwise mask-overlap fraction $S[i,j]/K$ averaged over off-diagonal entries and at its largest off-diagonal pair. $S[i,j]$ is the count of neurons active in both task $i$'s and task $j$'s stored mask; $K$ is the per-task budget. The chance baseline for uniformly random $K$-of-$H$ subsets is $K/H$ with $H=8192$. Across the AnnealSTE configurations, overlap is well above chance, confirming that the configurer reuses neurons rather than reducing to fixed disjoint subnetworks. The NoMask and FixedMask baselines bracket the range at 1.0 and 0.0.

\begin{table}[ht]
\centering
\small
\caption{Pairwise mask-overlap fraction $S[i,j]/K$ across configurations. ``mean off-diag'' averages over all $i\ne j$ entries; ``max off-diag'' is the largest pair per run, then averaged over seeds. Chance is the expected overlap for uniformly random $K$-of-$H$ subsets.}
\begin{tabular}{l c c c c c}
\toprule
Condition & $n$ & $K$ & mean off-diag & max off-diag & chance ($K/H$) \\
\midrule
\multicolumn{6}{l}{\textit{Task structure}} \\
\quad Correlated & 8 & 512 & $0.387 \pm 0.052$ & $0.577 \pm 0.022$ & $0.062$ \\
\quad Orthogonal & 8 & 512 & $0.365 \pm 0.025$ & $0.554 \pm 0.025$ & $0.062$ \\
\midrule
\multicolumn{6}{l}{\textit{Reconfig freq}} \\
\quad per-batch & 8 & 512 & $0.390 \pm 0.050$ & $0.585 \pm 0.031$ & $0.062$ \\
\quad per-task & 8 & 512 & $0.142 \pm 0.034$ & $0.313 \pm 0.069$ & $0.062$ \\
\midrule
\multicolumn{6}{l}{\textit{Budget K}} \\
\quad K=512 & 8 & 512 & $0.391 \pm 0.061$ & $0.589 \pm 0.042$ & $0.062$ \\
\quad K=128 & 8 & 128 & $0.424 \pm 0.034$ & $0.615 \pm 0.041$ & $0.016$ \\
\quad K=32 & 8 & 32 & $0.437 \pm 0.020$ & $0.688 \pm 0.022$ & $0.004$ \\
\quad K=8 & 8 & 8 & $0.346 \pm 0.042$ & $0.812 \pm 0.088$ & $0.001$ \\
\midrule
\multicolumn{6}{l}{\textit{Reconfig batch}} \\
\quad n=256 & 8 & 512 & $0.386 \pm 0.057$ & $0.579 \pm 0.045$ & $0.062$ \\
\quad n=64 & 8 & 512 & $0.413 \pm 0.020$ & $0.567 \pm 0.034$ & $0.062$ \\
\quad n=16 & 8 & 512 & $0.424 \pm 0.023$ & $0.595 \pm 0.036$ & $0.062$ \\
\quad n=4 & 8 & 512 & $0.397 \pm 0.014$ & $0.568 \pm 0.037$ & $0.062$ \\
\midrule
\multicolumn{6}{l}{\textit{MNIST}} \\
\quad default & 8 & 512 & $0.501 \pm 0.029$ & $0.665 \pm 0.045$ & $0.062$ \\
\midrule
\multicolumn{6}{l}{\textit{Baselines (synth)}} \\
\quad NoMask & 8 & 512 & $1.000 \pm 0.000$ & $1.000 \pm 0.000$ & $0.062$ \\
\quad FixedMask & 8 & 512 & $0.000 \pm 0.000$ & $0.000 \pm 0.000$ & $0.062$ \\
\midrule
\multicolumn{6}{l}{\textit{Baselines (MNIST)}} \\
\quad NoMask & 8 & 512 & $1.000 \pm 0.000$ & $1.000 \pm 0.000$ & $0.062$ \\
\quad FixedMask & 8 & 512 & $0.000 \pm 0.000$ & $0.000 \pm 0.000$ & $0.062$ \\
\bottomrule
\end{tabular}

\label{tab:mask_overlap}
\end{table}

\end{document}